\documentclass[12pt]{article}

\usepackage{latexsym,enumerate,bm,amsmath,amssymb}
\usepackage[dvips]{graphicx,color}
\usepackage{multirow}
\usepackage[latin1]{inputenc}
\newtheorem {Prop}{Proposition}
 
\newtheorem{Theo}{Theorem}
\newtheorem{Lemm}{Lemma}

\newcommand{\suc}{\mbox{$\{X_{n}\}$}}
\newcommand{\conv}{\rightarrow}

\DeclareRobustCommand{\FIN}{%
  \ifmmode 
  \else \leavevmode\unskip\penalty9999 \hbox{}\nobreak\hfill
  \fi
  $\bullet$ \vspace{5mm}}

\title{Supervised classification for a family of Gaussian functional models}

\author{Amparo Ba\'{\i}llo\footnote{These authors have been partially supported by Spanish grant MTM2007-66632.},   Juan Antonio Cuesta-Albertos\footnote{This author have been  partially supported by the Spanish  grant MTM2008-0607-C02-02.\newline
E-mail addresses: amparo.baillo@uam.es,  cuestaj@unican.es, antonio.cuevas@uam.es} and Antonio Cuevas$^*$\\
{\normalsize \it $^*$Universidad Aut\'onoma de Madrid and $^\dagger$Universidad de Cantabria }}
\date{}
\begin{document}

\maketitle

\begin{abstract}

In the framework of supervised classification (discrimination) for functional data, it is shown that the optimal classification rule can be explicitly obtained for a class of Gaussian processes with ``triangular'' covariance functions. This explicit knowledge has two practical consequences. First, the consistency of the well-known nearest neighbors classifier (which is not guaranteed in the problems with functional data) is established for the indicated class of processes. Second, and more important,
parametric and nonparametric plug-in classifiers can be obtained by estimating the unknown elements in the optimal rule.

The performance of these new plug-in classifiers is checked, with positive results, through a simulation study and a real data example.


\end{abstract}

\section{Introduction} \label{Section1}


\noindent \it Statement of the problem. Notation\rm

Discrimination, also called ``supervised classification'' in modern terminology, is one of the oldest statistical problems in
experimental science: the aim is to decide whether a random  observation $X$ (taking values in a ``feature space'' ${\mathcal F}$ endowed with a distance $D$)  either belongs to the population $P_0$ or to $P_1$.
For example, in a medical problem $P_0$ and $P_1$ could correspond to the group of ``healthy''
and ``ill'' individuals, respectively. The decision must be taken from the
information provided by a ``training sample'' $\mathcal X_n = \{ (X_i,Y_i), 1\leq i\leq n \}$.
Here $X_i$, $i=1,\ldots,n$, are independent replications of $X$, measured on $n$ randomly chosen
individuals, and $Y_i$ are the corresponding values of an
indicator variable which takes values 0 or 1 according to
the membership of the $i$-th individual to $P_0$ or $P_1$.
The term ``supervised'' refers to the fact that the individuals in the training
sample are supposed to be correctly classified, typically using ``external'' non
statistical procedures, so that they provide a reliable basis for the assignation
of the new observation. It is possible to consider the case where $K>2$ populations, $P_0,\ldots,P_{K-1}$ are involved but, in what follows, we will restrict ourselves to the binary case $K=2$.

The mathematical problem is to find a ``classifier'' (or ``classification rule'')
$g_n(x)=g_n(x;\mathcal X_n)$, with $g_n:{\mathcal F}\rightarrow
\{0,1\}$, that minimizes the classification  error ${\mathbb P}\{g_n(X)\neq Y\}$.
It is not difficult to prove (e.g., Devroye {\em et al}., 1996, p. 11) that the optimal classification rule
(often called ``Bayes rule'') is
\begin{equation} \label{opt}
g^*(x)={\mathbb I}_{\{\eta(x)>1/2\}}(x),
\end{equation}
where $\eta(x)={\mathbb E}(Y|X=x)$ and ${\mathbb I}_A$ stands for
the indicator function of a set $A\subset {\mathcal F}$. Of course, since $\eta$ is unknown the exact
expression of this rule is usually unknown, and thus
different procedures have been proposed to approximate $g^*$ using the training data.

From now on we will use the following notation. Let $\mu_i$ be the distribution of $X$
conditional on $Y=i$, that is,
$\mu_i(B) = {\mathbb P} \{ X\in B|Y=i \}$ for $B\in \mathcal B_{\mathcal F}$ (the Borel $\sigma$-algebra on ${\mathcal F}$) and $i=0,1$.
We denote by $S_i \subset \mathcal F$ the support of $\mu_i$, for $i=0,1$, $S=S_0\cap S_1$ and
$p={\mathbb P}\{Y=0\}$ (we assume $0<p<1$).
Given two measures $\mu$ and $\nu$, the expression $\mu << \nu$ denotes that $\mu$ is
absolutely continuous with respect to $\nu$ (i.e., $\nu(B)=0$ implies $\mu(B)=0$).

The notation $C[0,1]$ stands for the space of real continuous functions on the interval $[0,1]$ endowed 
with the usual supremum norm, denoted by $\Vert\cdot\Vert$.
The subspace of functions of class 2 (i.e. with two continuous derivatives)
is denoted by $C^2[0,1]$.

\

\noindent \it Finite dimensional spaces. Three classical discrimination procedures\rm

The origin of the discrimination problem goes back to the classical work by Fisher (1936) where, in the
$d$-variate framework ${\mathcal F}={\mathbb R}^d$, a simple ``linear classifier'' of type
$g_n(x)={\mathbb I}_{\{w^\prime x+w_0>0\}}$ was introduced for the case that both populations $P_0$ and $P_1$
are homoscedastic, that is, have a common covariance matrix $\Sigma$. Intuitively, $w^\prime x+w_0=0$ is chosen 
as the affine hyperplane which provides the ``maximum separation'' between both populations. It is well-known 
(see, e.g., Duda \it et al\rm. 2000 for details) that the the expression of Fisher's rule turns out to depend on the inverse $\Sigma^{-1}$ of the covariance matrix. It is also known that
Fisher's linear rule is in fact the optimal one (\ref{opt}) when the conditional distributions of $X|Y=0$ and
$X|Y=1$ are homoscedastic normals  and all the means and covariances are known. These conditions look quite restrictive but,
as argued by Hand (2006) in a provocative paper, Fisher's rule (or rather its sampling approximation obtained by estimating the unknown parameters) is hard to beat in practical examples. That is, while it is not difficult to construct examples where this rule outrageously fails, its performance is quite good in most cases found in real-life examples. For this reason, Fisher's linear rule is still the most popular classification tool among practitioners, in spite of the posterior intensive research on this topic. Thus, in a way, Fisher's rule represents a sort of ``golden standard'' in the multivariate statistical discrimination problem.


The books by Devroye \it et al\rm. (1996), Duda \it et al\rm. (2000) and Hastie \it et al\rm. (2001) offer different interesting perspectives of the work done in discrimination theory since Fisher's pioneering paper. All of them focus on the standard multivariate case ${\mathcal F}={\mathbb R}^d$. Many classifiers have been proposed as an alternative to Fisher's linear rule in this finite-dimensional setup. One of the simplest and easiest  to motivate is the so-called $k$-nearest neighbors method. Fixed a positive integer value (or smoothing parameter) $k=k_n$ this rule simply classifies an incoming observation $x$
in the population $P_1$ if the majority among the $k$ training observations closest to $x$ (with respect to the considered distance $D$) belong to $P_1$.
More concretely the $k$-NN rule can be defined by 
\begin{equation}
g_n(x) = {\mathbb I}_{ \{ \eta_n(x) > 1/2 \}},\label{kNNrule}
\end{equation}
 where
\begin{equation}
\eta_n(x) = \frac{1}{k} \sum_{i=1}^n {\mathbb I}_{\{ X_i\in k(x) \}} Y_i\label{kNNreg}
\end{equation}
and  ``$X_i\in k(x)$'' means that
$X_i$ is one of the $k$ nearest neighbors of $x$.

In fact, the definition of the $k$-NN rule is extremely simple and can be introduced (in terms of ``majority vote among the neighbors'') with no explicit reference to any regression estimator. However, the idea of replacing
the unknown regression function $\eta(x)$ in the optimal classifier
(\ref{opt}) with a regression estimator (given by (\ref{kNNreg}) in the case of the $k$-NN rule) is very natural. It suggests
a general methodology to construct a wide class of classifiers by just plugging in different regression estimators $\eta_n$ in (\ref{opt}) instead of the true regression function $\eta(x)$. In the finite dimensional case ${\mathcal F}={\mathbb R}^d$ this is a particularly fruitful idea, as a wealth of different (parametric and nonparametric) estimators of $\eta(x)$ is available; see Audibert and Tsybakov (2007) for some reasons in favor of the plug-in
methodology in classification. The main purpose of this work is to show that the \it plug-in methodology\/ \rm can be also successfully used for classification in some functional data models.

\

\noindent \it Discrimination of functional data. Differences with the finite-dimensional case\rm

We are concerned here with the problem of (binary) supervised classification with
functional data. That is,
we assume throughout that the space $({\cal F},D)$ where the
data $X_i$ live is a separable metric space (typically a space of functions).
For some theoretical results, considered below, we will impose more specific assumptions on ${\cal F}$.

The study of discrimination techniques with functional data
is not as developed as the corresponding finite-dimensional theory
but, clearly, is one of the most active research topics in
the booming field of functional data analysis (FDA).
Two well-known books including broad overviews of FDA with interesting examples are Ferraty
and Vieu (2006) and Ramsay and Silverman (2005). A recent survey on supervised and unsupervised classification with functional data can be found in Ba\'{\i}llo {\em et al}. (2009).

 While the formal statement of the functional classification problem is very much the same as that indicated at the beginning of this section, there are some important differences with the classical finite-\-dimensional case.

\begin{itemize}
\item[(a)] \it Lack of a simple functional version of Fisher's linear rule\rm: As mentioned above, the idea behind Fisher's rule requires to invert the covariance operator. When ${\mathcal F}={\mathbb R}^d$ this is increasingly difficult as the dimension $d$ increases, but it becomes impossible in the functional framework where the operator is typically not invertible. Thus the applicability of Fisher's linear methodology to functional data is a non-trivial issue of current interest for research.
See, for instance, James and Hastie (2001) and Shin (2008) for interesting
adaptations of linear discrimination ideas to a functional
setting.
\item[(b)] \it Difficulty to implement the plug-in idea\rm: Unlike the finite-dimensional case, the plug-in methodology is not generally considered as a standard procedure to construct functional classifiers. When $x$ is infinite-dimensional, there are yet few simple parametric models giving a good fit to the regression function and the structure of nonparametric estimators of $\eta$ is relatively complicated.
\item[(c)] \it The $k$-NN functional classifier is not universally consistent\rm:  In the discrimination problem a sequence of classifiers $\{g_n\}$, based on samples of size $n$, is said to be ``consistent'' when the corresponding sequence of classification errors converges, as $n$ tends to infinity, to the ``lowest possible error'' attained by the Bayes classifier (\ref{opt}); see Section \ref{Section3} below for more details. It turns out (see Stone, 1977) that, in the case of finite-dimensional data $X_i\in{\mathbb R}^d$,  any sequence of $k$-NN classifiers is consistent provided that $k_n\to\infty$ and $k_n/n\to 0$. Since such consistency holds  irrespectively of the distribution of the data $(X,Y)$, this property is called ``universal consistency''.

    The definition of the $k$-NN classifier can be easily translated to the functional setup (by replacing the usual Euclidean distance in ${\mathbb R}^d$ with an appropriate functional metric $D$). However, the universal consistency is lost. C\'erou and Guyader (2006, Th. 2) have obtained sufficient conditions for consistency of the  $k$-NN classifier when $X$ takes values in a separable metric space. Nevertheless, the required  assumptions are not always trivial to check. As the $k$-NN rule is a natural ``default choice'' in infinite-dimensional setups, an important issue is to ensure its consistency, at least for some functional models of practical interest.
\end{itemize}

\noindent \it The purpose and structure of this paper\rm

This work aims to partially fill the gaps pointed out in the points (b) and (c) of the above paragraph. 
To this end, in Subsection \ref{Subsection2_1} a simple expression is obtained for the Bayes (optimal) 
rule $g^*$ in the case that both distributions, $\mu_0$ and $\mu_1$, are equivalent. 
However, $g^*$ turns out to depend on the Radon-Nikodym derivative $d\mu_0/d\mu_1$ which is usually unknown, 
or has an extremely involved expression, even when  $\mu_0$ and $\mu_1$ are completely known. 
An interesting exception is given by Gaussian processes with a specific type of covariance functions, 
called ``triangular''. For these processes the Radon-Nikodym derivative has been explicitly calculated 
by Varberg (1961) and J\o rsboe (1968) whose results are collected and briefly commented in 
Subsection \ref{Subsection2_2}. In Subsection \ref{Subsection2_3} parametric plug-in estimators for $g^*$ 
are obtained by assuming that $\mu_0$ and $\mu_1$ are either
(parametric) Brownian motions or Ornstein-Uhlenbeck processes. 
Non-parametric plug-in estimators for $g^*$ are proposed and analyzed in Subsection \ref{Subsection2_4b}, 
under the sole assumption that the covariance functions are triangular. Since the proofs of the results in this subsection are rather technical, they are deferred to a final appendix. 
This concludes our contributions regarding issue (b). 
Section 3 is devoted to the  $k$-NN consistency problem introduced in (c): we use the above-mentioned 
result by C\'erou and Guyader (2006) to show that the $k$-NN rule is consistent in functional 
classification problems where the data are generated by certain Gaussian triangular processes specified 
in Subsection \ref{Subsection2_2}.

Finally, in Section \ref{Section4} the practical performance of the plug-in rules proposed in Section \ref{Section2} is checked, and compared with the $k$-NN rule,  through a simulation study and the analysis of a real data example.

\section{The optimal classifier for a Gaussian family} \label{Section2}

    \subsection{A general expression based on Radon-Nikodym derivatives} \label{Subsection2_1}


When the distributions $\mu_0$ and $\mu_1$ of $P_0$ and $P_1$ are both absolutely continuous with respect to some common $\sigma$-finite measure $\mu$, it is easy to see, as a consequence of Bayes formula, that the optimal rule is
\begin{equation}
g^*(x)={\mathbb I}_{\{(1-p)f_1(x)>pf_0(x)\}},\label{f0f1}
\end{equation}
 where $p={\mathbb P}\{Y=0\}$ and $f_0$, $f_1$ are the $\mu$-densities
of $P_0$ and $P_1$, respectively.

The expression (\ref{f0f1}) is particularly important in the finite dimensional problems with ${\mathcal F}={\mathbb R}^d$, where the Lebesgue measure $\mu$ arises as the natural reference measure and the corresponding Lebesgue densities can be estimated in many ways. In the infinite-dimensional spaces there is no such obvious dominant measure. However if we assume that $\mu_0$ and $\mu_1$, with supports $S_0$ and $S_1$, are absolutely continuous with respect to each other on $S_0\cap S_1$, the optimal rule can be also expressed in a simple way with respect to the Radon-Nikodym derivative ${d\mu_0}/{d\mu_1}$ as shown in the following result.

\begin{Theo} \label{Lareglaoptima}
Assume that $\mu_0 << \mu_1$ and $\mu_1 << \mu_0$ on $S=S_0\cap S_1$. Then
\begin{eqnarray}
\eta(x) & = & \left\{ \begin{array}{l}
0 \quad \mbox{if } x\in S_0\cap S^c
\\
1 \quad \mbox{if } x\in S_1\cap S^c
\\
\displaystyle \frac{1-p}{p \frac{d\mu_0}{d\mu_1}(x) + 1-p} \quad \mbox{if } x\in S .
\end{array} \right.
\label{EqLareglaoptima}
\end{eqnarray}
provides the expression for the optimal rule
$g^*(x) = {\mathbb I}_{ \{ \eta(x)> 1/2 \}}$.
\end{Theo}

\noindent {\sc Proof:} Define $\mu=\mu_0+\mu_1$. Then $\mu_i << \mu$,
for $i=0,1$, and we can define the Radon-Nikodym derivatives $f_i = d\mu_i/d\mu$, for $i=0,1$.
From the definition of the conditional expectation we know
that $\eta(x)=E(Y|X=x)=P(Y=1|X=x)$ can be expressed by
\begin{equation} \label{etaBayes}
\eta(x) = \frac{f_1(x)(1-p)}{f_0(x) p + f_1(x)(1-p)}.
\end{equation}
Observe that
$\mu \lvert_{S^c\cap S_i} = \mu_i\lvert_{S^c\cap S_i}$ and thus
$f_i \lvert_{S^c\cap S_i} = {\mathbb I}_{S^c\cap S_i}$, for $i=0,1$.
Since $\mu_0 << \mu_1$ and $\mu_1 << \mu_0$ on $S$ then, on this set, there exists the
Radon-Nikodym derivatives $d\mu_0/d\mu_1$ and $d\mu_1/d\mu_0$. In this case, it also holds
that $\mu\lvert_S << \mu_i\lvert_S$, for both $i=0,1$ and
$$
\frac{d\mu}{d\mu_i}(x) = 1 + \frac{d\mu_{1-i}}{d\mu_i} (x),
\qquad \mbox{for any } x\in S.
$$
Then (see, e.g., Folland 1999), for $i=0,1$ and for $P_X$-a.e. $x\in S$,
\begin{equation} \label{DRN}
f_i(x) = \frac{d\mu_i}{d\mu}(x) = \left( \frac{d\mu}{d\mu_i}(x) \right)^{-1}
    = \frac{1}{1 + \frac{d\mu_{1-i}}{d\mu_i} (x) }
\end{equation}
Substituting (\ref{DRN}) into expression (\ref{etaBayes}) we get (\ref{EqLareglaoptima}). \hfill{$\Box$}
\vspace{3 mm}

The mutual absolute continuity is not a very restrictive assumption if we deal with Gaussian measures. According to a well-known result by Feldman and H\'ajek (see Feldman, 1958)
for any given pair of Gaussian processes, there is a dichotomy in such
a way that they are either equivalent or mutually singular.
In the first case both measures $\mu_0$ and $\mu_1$ have a common support $S$. As for the identification of the support,
Vakhania (1975) has proved that if a Gaussian process, with trajectories in a
separable Banach space ${\cal F}$, is not degenerate
(i.e., the distribution of any non-trivial linear continuous functional is not degenerate)
then the support of such process is the whole space ${\cal F}$.

In any case, expression  (\ref{EqLareglaoptima}) would be of no practical use unless some
expressions, reasonably easy to estimate, can be found for the
Radon-Nikodym derivative $d\mu_0/d\mu_1$. This issue is considered in the next subsection.


\subsection{Explicit expression for a family of Gaussian distributions} \label{Subsection2_2}

The best known Gaussian process is perhaps the standard Brownian motion $\{W(t),\ t\geq 0\}$, for which
${\mathbb E}(W(t))=0$ and the covariance function is $\mbox{Cov}(W(s),W(t)):=\Gamma(s,t)=\min(s,t)$. A wide class
of Brownian-type processes can be obtained by location and scale changes of type $m(t)+\sigma W(t)$, where $m(t)$ 
is a given mean function and $\sigma>0$. 

In fact, the covariance structure $\Gamma(s,t)=\min(s,t)$
can be generalized to define a much broader class of processes with
$\Gamma(s,t) = u(\min(s,t)) \, v(\max(s,t))$, where $u$ and $v$ denote suitable real functions.
Covariance functions of this type are called \it triangular\rm. They have received considerable attention in the literature.
For example, Sacks and Ylvisaker (1966) use this condition in the study of optimal
designs for regression problems where the errors are generated by a zero
mean process with covariance function $\Gamma(s,t)$. It turns out that the Hilbert space
with reproducing kernel $K$ plays an important role in the results and, as these authors
point out, the norm of this space is particularly easy to handle when $\Gamma$ is triangular.
On the other hand, Varberg (1964) has given an interesting representation of the
processes $X(t),\ 0\leq t<b$, with zero mean and triangular covariance function. This author proved 
that they can be expressed in the form $X(t)=\int_0^bW(u)\,d_uR(t,u)$,
where $W$ is the standard Wiener process and $R=R(t,u)$ is a function, of bounded
variation with respect to $u$, defined in terms of $\Gamma$.

The so-called Ornstein-Uhlenbeck model, for which $\Gamma(s,t) = \sigma^2 \exp(-\beta|s-t|)$ ($\beta,\sigma>0$), provides another important class of processes with triangular
covariance functions. They are widely used in physics and finance.

The following theorem is due to Varberg (1961, Th. 1) and J\o rsboe (1968, p. 61). 
It shows that the Radon-Nikodym derivative can be expressed in a closed, relatively 
simple way for these special classes of Gaussian processes. 
For more information concerning explicit expressions of Radon-Nikodym derivatives 
for Gaussian processes see Segall and Kailath (1975) and references therein.  
From now on let us denote $m_i(t)={\mathbb E}\left(X(t)|Y=i\right)$.

\begin{Theo}\label{TheoExpresion}
\sl
Let $(\mathcal F,D) = (C[0,1],\| \cdot \|)$. Assume that $X|Y=i$, for $i=0,1$,
are Gaussian processes on $[0,1]$,  with covariance functions $\Gamma_i(s,t) = u_i(\min(s,t)) \, v_i(\max(s,t))$,
for $s,t\in[0,1]$, where $u_i,v_i$, for $i=0,1$, are positive functions in $C^{2}[0,1]$.
Assume also that $v_i$, for $i=0,1$, and $v_1u_1'-u_1v_1'$ are bounded away from zero on $[0,1]$,
that $u_1v_1'-u_1'v_1 = u_0v_0'-u_0'v_0$ and that $u_1(0)=0$ if and only if $u_0(0)=0$.
\begin{enumerate}[{\bf a)}]
\item
Assume that $m_i\equiv 0$, for $i=0,1$.
Then there exist some constants $C_1,C_2,C_3$ and a function $F$, whose expressions are given in the proof,
such that
\begin{equation}
\frac{d\mu_0}{d\mu_1}(x) = C_1 \exp \left[ \frac{1}{2} \left( C_3x^2(0) + C_2 x^2(1)
 - \int_0^1 \frac{x^2(t)}{v_0(t)v_1(t)} dF(t)\right) \right].
 \label{partea}
\end{equation}
\item Assume now that the covariance functions are identical, i.e. $u_i=u$ and $v_i=v$ for $i=0,1$, that $m_1\equiv 0$, $m_0$ is a function $m\in C^2[0,1]$,
such that $m(0)=0$ whenever $u(0)=0$. Then there exist some constants $D_1,D_2$ and a function $G$, whose expressions are given in the proof,
such that
\begin{equation}
\frac{d\mu_0}{d\mu_1}(x) = \exp \left\{ D_1 + \left( D_2 -2\,\frac{G(0)}{v(0)} \right) x(0)
    + 2 \,\frac{G(1)}{v(1)}\, x(1) - 2 \int_0^1 \frac{x(t)}{v(t)}\, dG(t) \right\}.\label{parteb}
\end{equation}
\end{enumerate}
\end{Theo}
\rm
\vspace{3 mm}

\noindent
{\sc Proof: }
\begin{enumerate}
\item[{\bf a)}] Varberg (1961, Th. 1) shows that, under the assumptions of (a),
$\mu_0$ and $\mu_1$ are equivalent measures. The Radon-Nikodym derivative of
$\mu_0$ with respect to $\mu_1$ is 
\begin{equation} \label{J1}
\frac{d\mu_0}{d\mu_1}(x) = C_1 \, \exp\left\{ \frac{1}{2} \left[ C_4 x^2(0) +
\int_0^1 F(t) d\left( \frac{x^2(t)}{v_0(t)v_1(t)} \right) \right] \right\},
\end{equation}
where
$$
C_1 = \left\{ \begin{array}{l}
\left( \frac{v_0(0)v_1(1)}{v_0(1)v_1(0)} \right)^{1/2} \; \mbox{if } u_0(0)=0 \\
\left( \frac{u_1(0)v_1(1)}{v_0(1)u_0(0)} \right)^{1/2} \; \mbox{if } u_0(0)\ne 0
\end{array} \right.
\quad
C_4 = \left\{ \begin{array}{l}
0 \; \mbox{if } u_0(0)=0 \\
\left( \frac{v_0(0)u_0(0)-u_1(0)v_1(0)}{v_1(0)v_0(0)u_0(0)u_1(0)} \right)^{1/2} \; \mbox{if } u_0(0)\ne 0
\end{array} \right.
$$
and $F=(v_1v_0'-v_0v_1')/(v_1u_1'-u_1v_1')$.

Observe that, by the assumptions of the theorem, $F$ is differentiable with bounded derivative.
Thus $F$ is of bounded variation and it may be expressed as the difference
of two bounded positive increasing functions.
Therefore the stochastic integral (\ref{J1}) is
well defined and it can be evaluated integrating by parts, leading to conclusion (\ref{partea}), with
$ C_3=C_4-F(0)/v_0(0)v_1(0) $ and $ C_2 = F(1)/v_0(1)v_1(1) $.

\item[{\bf b)}] In J\o rsboe (1968), p. 61, it is proved that, under the indicated assumptions, $\mu_0$ and $\mu_1$ are
equivalent measures with Radon-Nikodym derivative
$$
\frac{d\mu_0}{d\mu_1}(x) = \exp \left\{ D_3 + D_2 \, x(0) + \frac{1}{2} \int_0^1 G(t)
d\left( \frac{2x(t)-m(t)}{v(t)} \right) \right\},
$$
with
$$
D_3 = -\frac{m^2(0)}{2 \, u(0) \, v(0)} \mathbb I_{\{ u(0)>0 \}} , \quad
D_2 = \frac{m(0)}{u(0) \, v(0)} \mathbb I_{\{ u(0)>0 \}}\
$$
and $G=(vm'-mv')/(vu'-uv')$.
Again, the integration  by parts gives (\ref{parteb}), where $D_1 = D_3 - \int_0^1 G \, d(m/v)$.
\hfill{$\Box$}
\end{enumerate}
\vspace{3 mm}

In the general case where $m_0\neq m_1$ and $\Gamma_0\neq\Gamma_1$, let us denote by $P_{m,\Gamma}$ the distribution of the Gaussian process with mean $m$
and covariance function $\Gamma$. Then, applying the chain rule for Radon-Nikodym derivatives (see, e.g., Folland, 1999) we get 
\begin{equation} \label{RNNS}
\frac{d\mu_0}{d\mu_1}(x) = \frac{dP_{m_0,\Gamma_0}}{dP_{m_1,\Gamma_1}} (x)
    = \frac{dP_{m_0,\Gamma_0}}{dP_{0,\Gamma_0}} (x) \,
      \frac{dP_{0,\Gamma_0}}{dP_{0,\Gamma_1}} (x) \,
      \frac{dP_{0,\Gamma_1}}{dP_{m_1,\Gamma_1}} (x).
\end{equation}
Under the appropriate assumptions the expressions of the Radon-Nikodym derivatives
in the right-hand side of (\ref{RNNS}) are given in (\ref{partea}) and (\ref{parteb}).

\subsection{Parametric plug-in rules} \label{Subsection2_3}



The aim of this subsection is twofold. First and foremost, we show how the theoretical results of Subsections 
\ref{Subsection2_1} and \ref{Subsection2_2} become useful in practice.  To this end, we consider examples of well-\-known Gaussian processes
that fulfill the requirements of Theorems \ref{Lareglaoptima} and \ref{TheoExpresion}, namely Brownian motions with drift and Ornstein-\-Uhlenbeck processes. We derive the expressions of the Radon-\-Nikodym derivatives $d\mu_0/d\mu_1$ for these examples. Then, it is straightforward to compute the Bayes rule $g^*$ for classification between two elements
of one of these families.
In these particular examples the mean and variance of the Gaussian process $X|Y=i$ have known
parametric expressions (up to a finite number of parameters).
Thus $g^*$ is completely specified as long as the parameters have known values.
When this is not the case, we can substitute each unknown parameter in $g^*$ by some estimate.
The resulting discrimination procedure is called the {\em parametric plug-in} rule.
In particular, for the Bayes rules given in (\ref{BrownianBayes}), (\ref{BayesRuleBrRn}), (\ref{OUBayes})
and (\ref{RNDerOURandAt0}) below the explicit expression of the parameter estimates is given in the appendix. 

The second objective of Subsection \ref{Subsection2_3} is to obtain the expressions of the Bayes 
rules for the models used in Section \ref{Section4} and to derive the corresponding parametric plug-in versions.
\vspace{3 mm}

\noindent
{\em Two Brownian motions}

Let us denote $X(t;i)=(X(t)|Y=i)$. In the Brownian case, using the standard notation in stochastic differential equations, $X(t;i)$ is just the solution of    $dX(t;i) = m_i(t)\, dt + \sigma_i W_i(t) \, dt $, for $i=0,1$ and $t\in[0,1]$. Here
$m_1\equiv 0$, $m_0(t)=ct$, $0<c<\infty$ is a constant, $W_0$ and $W_1$ are two uncorrelated Brownian motions
and $(X(0;i)\sim N(0,\theta_i^2)$.
Then, if $\sigma_0=\sigma_1=\sigma$, the conditions of Theorem \ref{TheoExpresion} are satisfied
with $u_i(t) = \theta_i^2+\sigma^2 t$ and $v_i\equiv 1$, for $i=0,1$.

When $\theta_0=\theta_1=0$, we have $X(0;i)\equiv 0$ and, for any $x\in S$,
$$
\frac{d\mu_0}{d\mu_1}(x) = \exp \left\{ \frac{c}{\sigma^2}(2\, x(1)-c) \right\}.
$$
Thus the Bayes rule is
\begin{equation} \label{BrownianBayes}
g^*(x) = {\mathbb I}_{ \{x(1) < c/2 \} }.
\end{equation}
If $\theta_i\neq 0$ for $i=0,1$, then $X(0;i)$ is random and a similar calculation yields that
 the Bayes rule classifies $x$ in population $P_1$ whenever
\begin{equation} \label{BayesRuleBrRn}
\frac{c}{\sigma^2} \left[ 2 (x(1)-x(0)) -c \right]
+ \frac{1}{2} \left( \frac{1}{\theta_1^2}-\frac{1}{\theta_0^2} \right) x^2(0)
< \log\left(\frac{\theta_0}{\theta_1}\right).
\end{equation}
Replacing the unknown parameters, $c$, $\sigma$ and $\theta_i$ in (\ref{BrownianBayes}) and
(\ref{BayesRuleBrRn}) by estimates, we obtain the corresponding {\em parametric plug-in} rules.

When $\sigma_0\neq\sigma_1$, then $u_i(t) = \theta_i^2 + \sigma_i^2 t$, $v_i \equiv 1$, for $i=0,1$,
and the hypothesis $u_1v_1'-u_1'v_1=u_0v_0'-u_0'v_0$
in Theorem 2 is not satisfied.
In fact, if this last equality does not hold, by Theorem 1 in Varberg (1961) we know that $\mu_0$
and $\mu_1$ are mutually singular.
\vspace{3 mm}

\noindent
{\em Two Ornstein-Uhlenbeck processes}


Let $X|Y=i$, for $i=0,1$, be Ornstein-Uhlenbeck processes given by
$$
dX(t;i)= - \, \beta_i \, (X(t;i)-\eta_i) \, dt + \sqrt{2\beta_i} \, \sigma_i \, dW_i(t),
$$
where $W_0$ and $W_1$ are two independent Brownian motions and $\beta_i>0$, $\sigma_i>0$, $\eta_i$
are constants.

If $X(0;i)$ is equal to a constant $c_i$, we have that
$m_i(t) = \eta_i + (c_i-\eta_i)e^{-\beta_i t}$ and 
$\Gamma_i(s,t) = \sigma_i^2 \left(e^{-\beta_i|s-t|}-e^{-\beta_i|s+t|}\right)$.
Fixing $v_i(1)=1$, we get $u_i(t)=\sigma_i^2 e^{-\beta_i} (e^{\beta_i t}-e^{-\beta_i t})$
and $v_i(t)= e^{\beta_i(1-t)}$ for $i=0,1$.
The condition $u_1v_1'-u_1'v_1=u_0v_0'-u_0'v_0$ in Theorem \ref{TheoExpresion}
is fulfilled if and only if $\beta_0 \sigma_0^2=\beta_1 \sigma_1^2$.
Also, since $u_i(0)=0$, then $m_i(0)=c_i$ has to be 0 for $i=0,1$.
Then it is straightforward to check that the Bayes rule $g^*$ classifies $x$ in population $P_1$ if
\begin{eqnarray}
\lefteqn{0 > 2\left(\beta_0^2(\sigma_0^2-\eta_0^2)-\beta_1^2(\sigma_1^2-\eta_1^2)\right)
+ 4 \, x(1) ( \eta_0\beta_0 - \eta_1\beta_1 )
+ (\beta_1-\beta_0) \, x^2(1)}
 \nonumber \\
 & &  \hspace{3.5 cm} + \, 4 \, (\eta_0\beta_0^2-\eta_1\beta_1^2) \int_0^1 x(t) \, dt
     + (\beta_1^2-\beta_0^2) \int_0^1 x^2(t) \, dt . \label{OUBayes}
\end{eqnarray}

When $X(0;i)$ is random, it follows a normal distribution with mean $\eta_i$ and variance $\sigma_i^2$. Then
$m_i(t) = \eta_i$, for all $t\in[0,1]$, and $\Gamma_i(s,t)=\sigma_i^2e^{-\beta_i|s-t|}$,
$u_i(t)=\sigma_i^2e^{-\beta_i(1-t)}$ and $v_i(t)=e^{\beta_i(1-t)}$. Consequently,
the Bayes rule assigns $x$ to population $P_1$ if
\begin{eqnarray}
\lefteqn{2 \beta_1\sigma_1^2(\log(\beta_1)-\log(\beta_0))
> 2\left[\beta_0^2\sigma_0^2-\beta_1^2\sigma_1^2
     + \beta_1\eta_1^2(1+\beta_1)-\beta_0\eta_0^2(1+\beta_0)\right] } \nonumber \\
 & & \hspace{3.5 cm}
 + 4 \, x(1) ( \eta_0\beta_0 - \eta_1\beta_1 )
     + \, 4 \, (\eta_0\beta_0^2-\eta_1\beta_1^2) \int_0^1 x(t) \, dt \nonumber \\
 & & \hspace{3.5 cm}
 + (\beta_1-\beta_0) \left[ x^2(0) + x^2(1)
     + (\beta_1+\beta_0) \int_0^1 x^2(t) \, dt \right]  . \label{RNDerOURandAt0} 
\end{eqnarray}

The parametric plug-in classification rule is derived by substituting the unknown parameters
$\beta_i$, $\eta_i$ and $\sigma_i$, $i=0,1$, in (\ref{OUBayes}) and (\ref{RNDerOURandAt0}) with their corresponding
estimators. 


\subsection{Nonparametric plug-in rules} \label{Subsection2_4b}

In this section we analyze the situation in which the processes ultimately belong to the Gaussian family fulfilling the conditions of Theorem \ref{TheoExpresion},  but we do not place any parametric assumption on the mean and the covariance functions. However, let us note that, until we get to the estimation of the Radon-Nikodym derivatives, the Gaussianiaty assumption is not needed. Specifically, we only assume that the covariance functions of the involved processes are of type $\Gamma(s,t)=u(\min(s,t))v(\max(s,t))$, for some (unknown) real functions $u$, $v$ where $v$ is bounded away from 0 on the interval $[0,1]$.

Observe that, in order to use a plug-in version of the optimal classification rule along the lines of Theorems \ref{Lareglaoptima} and  \ref{TheoExpresion}, we need to estimate the functions $m$, $u$ and $v$ as well as their first and second derivatives. Since these estimation problems have some independent interest, in this subsection we consider them
in a general setup, not necessarily linked to the classification problem. Thus we use the ordinary iid sampling model with a fixed sample size denoted, for simplicity, by $n$ in all cases.

Regarding $u$ and $v$, let us note that the condition $\Gamma(s,t)=u(\min(s,t))v(\max(s,t))$, for $s,t\in[0,1]$,
entails $u(s)=\Gamma(s,1)/v(1)$ and $v(t)=\Gamma(0,t)/u(0)$ if $u(0)>0$. However, it is clear that these conditions only determine $u$ and $v$ up to multiplicative constants  so that one can impose (without loss of generality) the additional assumption $v(1)=1$. Thus, it turns out that $u$ and $v$ can be uniquely determined in terms of $\Gamma(0,t)$ and $\Gamma(s,1)$.
Our study will require three steps: first, the estimation of the mean function $m$ and its derivatives, then the analogous study for $\Gamma(0,t)$, $\Gamma(s,1)$ and $\sigma^2(t):=\Gamma(t,t)$ and, finally, the analysis of more involved functions defined in terms of these.

In Propositions 1 to 3 below we assume that the sample data are $X_1,\ldots,X_n$, iid trajectories of a process 
$X$ in the space $C [0,1]$, endowed with the supremum norm, $\| \cdot \|$.
\vspace{3 mm}

\noindent
{\it
Estimation of the mean and covariance functions and their derivatives}  

To estimate the mean function $m(t)={\mathbb E}\left[ X(t)\right]$ and its derivatives, we will only need to assume that \suc \ satisfies that ${\mathbb E}\|X_1\|^2<\infty$, which (see p. 172 in Araujo and Gin\'e, 1980) implies that the distribution of $X_1$ satisfies the Central Limit Theorem (CLT) in $(C [0,1],\| \cdot \|)$.

The natural estimator of $m$  is the sample mean, denoted by $\hat m_n (t)=\sum_{i=1}^n X_i(t)/n$. Since the derivatives of $m$ are also involved in the expressions
of the Radon-Nikodym derivatives obtained in Theorem \ref{TheoExpresion}, we will also need to consider the estimation of $m^\prime$ and $m^{\prime\prime}$. Our estimators will depend on  a given sequence $h_n\downarrow 0$  of smoothing parameters. Given $t \in [h_n,1-h_n]$, define
\begin{eqnarray*}
\hat m_n'(t) := \frac {\hat m_n(t+h_n)-\hat  m_n(t-h_n)}{2h_n},\qquad
\hat m_n''(t) := \frac {\hat m_n(t+h_n) +\hat  m_n(t-h_n)- 2 m_n(t)}{h_n^2}.
\end{eqnarray*}

For $t \in [0,h_n)$, we define
\begin{eqnarray*}
\hat m_n'(t) := \frac {\hat m_n(t+h_n)-\hat  m_n(0)}{h_n+t},
\qquad
\hat m_n''(t) := \frac {\hat m_n(t+h_n) +\hat  m_n(0)- 2\hat  m_n(\gamma_n)}{\gamma_n^2}.
\end{eqnarray*}
where $\gamma_n = (t+h_n)/2$. The definition of $\hat m_n'$ and $\hat m_n''$ on $(1-h_n,1]$ is similar.
These definitions allow us to handle analogously the extreme points and the inner ones. Thus we will not pay special attention to the extreme points in the proofs.

There is a slight notational abuse in these definitions as, for example, $\hat m_n'(t)$ is not the derivative of $\hat m_n(t)$ but an estimator of $m^\prime(t)$. We keep this notation throughout the manuscript for simplicity.

As mentioned at the beginning of this section, due to the triangular structure of $\Gamma$, in principle
we should only concentrate on the estimation of the functions $s\mapsto \Gamma(s,1)$ and $t \mapsto\Gamma ( 0,t)$ 
and their derivatives. 
However, due to technical reasons we will also need to consider the function $\sigma^2(t)=\Gamma(t,t)$ and its derivatives.
Natural nonparametric estimators of these functions can be given in terms of the empirical covariance
\[
\hat \Gamma_n(s,t):= \frac 1 n \sum_i \left(X_i(s)-\hat m_n (s)\right)  \left(X_i(t)-\hat m_n (t)\right),\ s,t\in[0,1].
\]
The estimation of the required derivatives is carried out in an analogous way as we did with the mean function.
Observe finally that, since $v(1)=1$, we can estimate $u(t)= \Gamma(t,1)$ by
$\hat u_n(t) := \hat \Gamma_n (t,1)$ for any $t \in [0,1]$
and similarly for its first two derivatives. Regarding the function $\sigma^2$, we estimate $\sigma^2(t)$  by
$\hat \sigma_n^2(t):=\hat \Gamma_n(t,t)$.

\begin{Prop} \label{TheoMedias}

Let  \suc \ be iid trajectories in $C [0,1]$ of a process such that ${\mathbb E}\|X_1\|^2<\infty$ and whose mean function $m: [0,1] \conv \mathbb R$ has a Lipschitz second derivative.

\begin{enumerate}[{\bf a)}]
\item For the mean estimation problem we have,
\begin{eqnarray}
\label{TheoRes1}
\| m - \hat m_n \| & = & O_P(n^{-1/2})
\\
\label{TheoRes2}
\| m' - \hat m_n' \| & = & O_P\left((n^{1/2}h_n)^{-1}\right) + O(h_n^2)
\\
\label{TheoRes3}
\| m'' - \hat m_n''\| &=& O_P\left((n^{1/2}h_n^2)^{-1}\right) +O(h_n)
\end{eqnarray}
\item Assume that ${\mathbb E}\|X_1\|^4<\infty$ and that the functions $t\conv \Gamma(t,1)$, $t \conv \Gamma ( 0,t)$ and $\sigma^2$ admit  Lipschitz second order derivatives. Then, we have
\begin{eqnarray}
&&\|\hat \Gamma_n( \cdot,1) - \Gamma(\cdot,1)\|  =\|\hat u_n - u\| =  O_P(n^{-1/2}),\label{EqThV.u}
\\
&&\|\hat \Gamma_n^\prime(\cdot,1) - \Gamma^{\prime}(\cdot,1)\| =  \|\hat u'_n - u'\|  = O_P\left(\left(n^{1/2}h_n\right)^{-1}\right)+ O(h_n^2),\label{EqThV.u1}
\\
&&\|\hat \Gamma_n^{\prime\prime}(\cdot,1) - \Gamma^{\prime\prime}(\cdot,1)\|  =\|\hat u''_n - u''\|  = O_P\left(\left(n^{1/2}h_n^2\right)^{-1}\right)+ O(h_n),\label{EqThV.u2}
\end{eqnarray}
Similar results also hold for $\hat \Gamma_n(0, \cdot)$ and $\hat \sigma^2_n$.

\end{enumerate}

\end{Prop}

From the proof of this proposition (see the Appendix) it can be checked that the assumption ${\mathbb E}\Vert X_1\Vert^4<\infty$
can be replaced with ${\mathbb E}\Vert X_1\Vert^{2+\delta}<\infty$, for some $\delta>0$, and ${\mathbb E}(X^r(1))<\infty$ for any $r>0$.
\vspace{3 mm}

\noindent
{\it
Estimation of $v$}

The estimation of $v$ is harder than that of $u$. It will be useful to distinguish two cases, 
where the estimators must be defined in different ways.
In the case $u(0)>0$ (corresponding to the case $\sigma^2(0)>0$) we have $ v(t)= \Gamma(0,t)/u(0)$ which is estimated by
\begin{equation}
\label{Defn_n1}
\hat v_n(t):=  \frac 1 {\hat u_n(0)} \hat \Gamma_n(0,t),  t \in [0, 1].
\end{equation}

When $u(0)=0$ (which implies that $\sigma^2(0)=0$), the estimator proposed  in (\ref{Defn_n1}) is, at best, highly unstable.
This case is not unusual: see, for instance, the examples introduced in  Subsection \ref{Subsection2_3}  when $X(0)/Y=i$ is constant.
For the sake of simplicity from now on assume that $\sigma^2(t)>0$ for $t \in (0,1)$.

The first step is to define $\hat v_n(t)=\hat \sigma_n^2(t)/\hat u_n(t)$ for $t\in[\delta_n,1]$, 
where $\delta_n$ is a sequence of positive numbers converging to zero (whose rate will be determined later). 
Then we define estimates for the first and the second derivatives of $v$ on the same interval. 
The structure of $v_n$ as a quotient suggests defining, on $[\delta_n,1]$,
\begin{eqnarray*}
\hat v_n'  &:=&
\frac{1}{\hat u^2_n}\left( (\hat \sigma^2_n)' \hat u_n - \hat u_n'\hat \sigma^2_n  \right),
\\
\hat v_n'' &:=&
\left.\left.\frac{1}{\hat u^3_n}
\right(\hat u_n \left( (\hat \sigma^2_n)'' \hat u_n - \hat u_n''\hat \sigma^2_n \right)
-
 2  \hat u_n'( (\hat \sigma^2_n)' \hat u_n - \hat u_n'\hat \sigma^2_n )
\right),
\end{eqnarray*}
where $(\hat \sigma^2_n)'(t)=\hat\Gamma_n^\prime(t,t)$, $(\hat \sigma^2_n)''(t)=\hat\Gamma_n^{\prime\prime}(t,t)$

Now we complete the definition of our estimator of $v$ on the whole interval by using a  Taylor-kind expansion on $[0,\delta_n)$,
\begin{equation}
\label{EqV_0_2}
\hat v_n^{}(t)  =
\hat v_n(\delta_n) + (t-\delta_n) \hat v_n'(\delta_n) + \frac 1 2  (t-\delta_n)^2 \hat v_n''(\delta_n),\quad  \mbox{ if } t \in [0, \delta_n).
\end{equation}
Finally, take
\begin{eqnarray}
\hat v_n'^{}(t)  &:=&
 \hat v_n'(\delta_n) +  (t-\delta_n) \hat v_n''(\delta_n),\quad \mbox{ if } t \in [0, \delta_n).
\nonumber \\
[2mm]
\hat v_n''^{}(t)  &:=&
\hat v_n''(\delta_n),{\phantom{+  (t-\delta_n) \hat v_n''^{(1)}(\delta_n)}} \mbox{ if } t \in [0, \delta_n).
\nonumber
\end{eqnarray}

\begin{Prop} \label{TheoFuncv}
Let the assumptions of Proposition \ref{TheoMedias} (b) hold.
\begin{enumerate}[{\bf a)}]
\item If $u(0)>0$ then the rate of convergence of $\Vert \hat v_n-v\Vert$, $\Vert \hat v_n^\prime-v^\prime\Vert$ and
    $\Vert \hat v_n^{\prime\prime}-v^{\prime\prime}\Vert$ are the same as those of (\ref{EqThV.u}), (\ref{EqThV.u1}) and (\ref{EqThV.u2}), respectively.
\item If $u(0)=0$  assume that $\inf_t u'(t)>0$ and $\inf_{t \in [\delta,1]} \sigma^2 (t) >0$ for every $\delta >0$.  Let $\{\delta_n\}\downarrow 0$ be such that $\sup (n^{-1/2},h_n) = o(\delta_n)$. Then
    \begin{eqnarray*}
\| \hat v_n - v \| & = & O_P\left(\frac{\delta_n}{h_n^2 \sqrt n}\right) + O(h_n) +O(\delta_n^3)
\\
\| \hat v_n' - v' \| & = & O_P\left(\frac{1}{h_n^2 \sqrt n}\right) + O\left(\frac{h_n}{\delta_n}\right) +O(\delta_n^2)
\\
\| \hat v_n'' - v'' \| & = & O_P\left(\frac{1}{\delta_n h_n^2 \sqrt n}\right) + O\left(\frac{h_n}{\delta_n^2}\right) + O(\delta_n).
\end{eqnarray*}
\end{enumerate}
\end{Prop}
\vspace{3 mm}

\noindent
{\it
Estimation of the Radon-Nikodym derivatives}

Here we plug-in the estimates of $m$, $u$, $v$ and their derivatives obtained above in the Radon-Nikodym derivatives
$f=d\mu_0/d\mu_1$ obtained above in Theorem \ref{TheoExpresion}. Denote by $\hat f_n$ the resulting estimate. Then, we compute  the convergence rate to the Bayes risk of the error attained by the corresponding nonparametric plug-in classification procedure.

According to Theorem \ref{TheoExpresion} the Radon-Nikodym densities of interest are the exponential of some integrals, ratios, products or square roots of  functions estimated with orders of convergence appearing  in Propositions \ref{TheoMedias} and \ref{TheoFuncv}.
The final rate will be that of the worst estimate handled, which corresponds to  the second order derivatives.
As with the estimation of $v$, there is some difference in the orders depending on whether $\sigma^2(0)$ is strictly positive or not.

The main conclusions are summarized in the following result.

\begin{Theo}\label{TheoRatesC}
Let us assume that conditions in Proposition \ref{TheoMedias} (b) and Theorem \ref{TheoExpresion} hold.
\begin{enumerate}[{\bf a)}]
\item If $u_i(0)>0$ for $i=0,1$, then for $h_n=O(n^{-1/6})$ we get
\[
\log {\hat f_n (x)} - \log \frac {d\mu_0}{d \mu_1}(x) = O_P\left({n^{-1/6}}\right), \ x \in {\mathcal C}[0,1].
\]

\item If $u_i(0)=0$ for $i=0,1$ and $\inf_t u'(t)>0$ and $\inf_{t \in [\delta,1]} \sigma^2 (t) >0$ for every $\delta >0$, then, for $h_n=O(n^{-9/50})$ we have
\[
{\mathbb E}\left(\left.\log \hat f_n (X) - \log \frac {d\mu_0}{d \mu_1}(X)\right|X_1,\ldots,X_n\right) = O_P\left({n^{-1/10}}\right).
\]
\end{enumerate}
\end{Theo}

Let us note that, in any case, our nonparametric estimator $\hat f_n(x)=dP_{\hat m_0\hat\Gamma_0}/dP_{\hat m_1\hat\Gamma_1}$ is constructed, using (\ref{RNNS}), under the sole assumption that the covariance function has a triangular structure. So, the estimator is formally the same in both cases a) and b) of Theorem \ref{TheoExpresion}.
If we knew that $m_i=0$ for $i=0,1$ then we could employ $\hat f_n(x)=dP_{\hat m_0\hat\Gamma_0}/dP_{\hat m_0\hat\Gamma_1}$ and the rates of Theorem \ref{TheoRatesC} would improve, under the assumptions of Theorem \ref{TheoRatesC} b), to $O_P({n^{-3/28}})$.
\vspace{3 mm}

\noindent {\it Using higher order derivatives}

The proof of Theorem \ref{TheoRatesC} was based on the use of Taylor expansions of order two. 
Next we show how the existence of higher order derivatives improves the estimation process.

\begin{Prop}\label{CoroTheoRatesC}
Under the assumptions of Theorem \ref{TheoRatesC} suppose further that the mean function 
$m: [0,1] \conv \mathbb R$ as well as the functions $t\conv \Gamma(t,1)$, $t \conv \Gamma ( 0,t)$ 
and $\sigma^2$ admit  Lipschitz third order derivatives. Then the rates in Theorem 3 a) and b) are 
improved to $O_P(n^{-1/4})$ and $O_P(n^{-5/32})$, respectively.
\end{Prop}

A remark similar to that made after Theorem \ref{TheoRatesC} applies here. If we incorporate the information
$m_i=0$ to the estimator, the convergence rate in Proposition \ref{CoroTheoRatesC} b) slightly improves to $O_P({n^{-1/6}})$.

The convergence orders may be further improved by assuming additional smoothness orders and taking advantage of numerical differentiation techniques (see, for instance, p. 146 in Gautschi, 1997). We will not develop this idea in the present work. However, let us observe that in the estimation of functions with infinite derivatives it is possible to obtain orders as close to $O_P(n^{-1/2})$ as desired by choosing $k$ large enough in the $k$-point rule (see, for instance, Herzeg and Cvetkovic, 1986).
\vspace{3 mm}

\noindent{\it Estimation of the probability of misclassification}

We denote by $\hat L_n:=L(\hat g_n)={\mathbb P}\{ \hat g_n(X)\neq Y|\mathcal X_n\}$ the classification error
associated with the nonparametric plug-in rule  $\hat g_n(x)={\mathbb I}_{\{\hat \eta_n(x)>1/2\}}$. Here
$\hat \eta_n$ is obtained by substituting
the Radon-Nikodym derivative $f=d\mu_0/d\mu_1$ in (\ref{EqLareglaoptima}) with the estimator $\hat f_n$
obtained by replacing $m$, $u$, $v$ and their derivatives with the corresponding nonparametric estimators obtained
along this subsection.
The following result is an example of how the convergence rates for the difference between the
logarithms of the Radon-Nikodym derivatives $\hat f_n(x)$ and $f(x)$ can be translated into  convergence rates of $\hat L_n$ to the Bayes error $L^*$.

\begin{Theo} \label{latasaparaelerror}
Let the assumptions of Proposition~\ref{TheoMedias} (b) and Theorem~\ref{TheoExpresion} hold.
If $u_i(0)>0$ for $i=0,1$, then taking $h_n=O(n^{-1/6})$ we get $\hat L_n-L^*=O_P\left({n^{-1/6}}\right)$.
 \end{Theo}

In the case when $u_i(0)=0$, for $i=0,1$, we can prove that $\hat L_n-L^*$ is $O_P(n^{-1/10})$
under the assumptions that $\inf_t u'(t)>0$ and $\inf_{t \in [\delta,1]} \sigma^2 (t) >0$ for every $\delta >0$.
The idea is to follow the same steps as in the proof of Theorem~\ref{latasaparaelerror}, but bounding the integrals in
(\ref{CondExpec1}) and (\ref{Integral2}) as we did along the proof of Theorem 3. 

\section{Consistency of the $k$-NN functional rules} \label{Section3}

As stated in the introduction, the $k$-NN classifier is not universally consistent in the functional setting. 
However, C\'erou and Guyader (2006) provide sufficient conditions for the consistency $L_n\rightarrow L^*$ 
in probability (or, equivalently, ${\mathbb E}(L_n)\rightarrow L^*$), where $L_n$ is the conditional 
classification error of the $k$-NN rule. In this section we show that these conditions are fulfilled by 
the Gaussian processes introduced in Section \ref{Subsection2_2} and, in consequence, that the $k$-NN 
is consistent in probability for them.

Throughout this section the feature space where the
variable $X$ takes values is a separable metric space $({\cal
F},D)$. As usual, we will denote by $P_X$ the distribution of $X$
defined by
$P_X (B) = {\mathbb P} \{  X\in B \} \quad \mbox{for } B\in\mathcal B_{\mathcal
 F}$, where $\mathcal B_{\mathcal F}$ are the Borel sets of $\mathcal F$.

The key assumption is a
regularity condition on the regression
function $\eta(x)={\mathbb E}(Y|X=x)$ which is called \it Besicovich condition\/ \rm {\bf (BC)}. The function $\eta$ is said to fulfill {\bf (BC)} if 
$$
\lim_{\delta\to 0} \frac{1}{P_X(B_{X,\delta})} \int_{B_{X,\delta}} \eta(z) \, dP_X(z) = \eta(X)
\quad \mbox{in probability},
$$
where $B_{x,\delta} := \{ z\in \mathcal F: D(x,z)\leq \delta \}$ is the closed ball
with center $x$ and radius $\delta$.
 Besicovich condition plays, for instance, an important role in the consistency of kernel rules
(see Abraham {\em et al}. 2006).

C\'erou and Guyader (2006, Th. 2) have proved that, if $({\cal F},D)$ is separable and condition \bf (BC)\/ \rm  is
fulfilled, then the $k$-NN classifier defined by (\ref{kNNrule}) and (\ref{kNNreg}) is consistent in probability
provided that $k_n \to\infty$ and $k_n /n\to 0$\rm.
In order to apply this result in our case, it will be sufficient to observe that the continuity
($P_X$-a.e.) of $\eta(x)$ implies also {\bf (BC)}. Consequently we can establish the following result, 
whose proof is immediate from Theorems \ref{Lareglaoptima} and \ref{TheoExpresion}.

\begin{Prop}
Under the assumptions of Theorem \ref{Lareglaoptima} suppose that $P_X(\partial S)=0$. Then
for $P_X$-a.e. $x,z$ in the topological interior of $S$,
\begin{equation}
\label{Lips}
|\eta(z)-\eta(x)|  = \left| \frac{1-p}{p \frac{d\mu_0}{d\mu_1}(z) + 1-p} -
\frac{1-p}{p \frac{d\mu_0}{d\mu_1}(x) + 1-p} \right|
\leq \frac{p}{1-p} \left|\frac{d\mu_0}{d\mu_1}(x) - \frac{d\mu_0}{d\mu_1}(z)\right|.
\end{equation}
As a consequence, for both cases a) and b) considered in Theorem \ref{TheoExpresion} the $k$-NN functional 
classifier is consistent in probability, provided that $k_n\to\infty$ and $k_n/n\to 0$.

\end{Prop}

Of course, the point is that the Radon-Nikodym derivatives given in Theorem \ref{TheoExpresion} 
are continuous on $C[0,1]$. So (\ref{Lips}) would imply also the continuity of $\eta(x)$ which in 
turn entails the Besicovich condition {\bf (BC)} and the consistency.


\section{Empirical results} \label{Section4}

In this section we compare the performance of the $k$-NN classification procedure with the
plug-in one for infinite-dimensional data. First (Subsection~\ref{Subsection4_1}) we describe the results of a simulation study
carried out with processes from the two Gaussian families specified in Subsection~\ref{Subsection2_3}.
Afterwards (Subsection~\ref{Subsection4_2}) we focus on a real-data set.

\subsection{Monte Carlo study} \label{Subsection4_1}

\begin{figure} 
\begin{center}
\begin{tabular}{cc}
\resizebox{7 cm}{5 cm}{\includegraphics*[4cm,9.4cm][17cm,20.3cm]{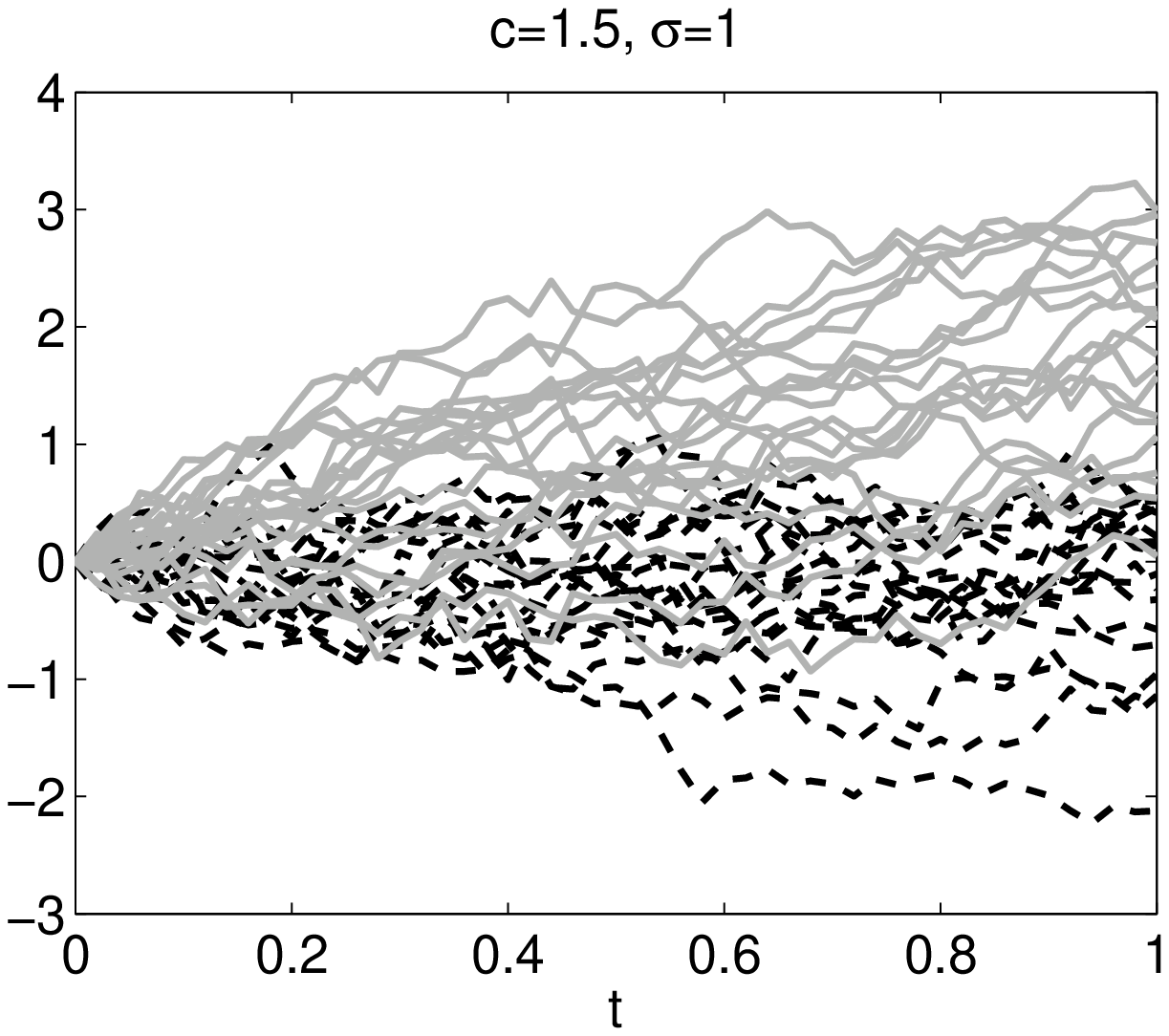}}
&
\resizebox{7 cm}{5 cm}{\includegraphics*[4cm,9.4cm][17cm,20.3cm]{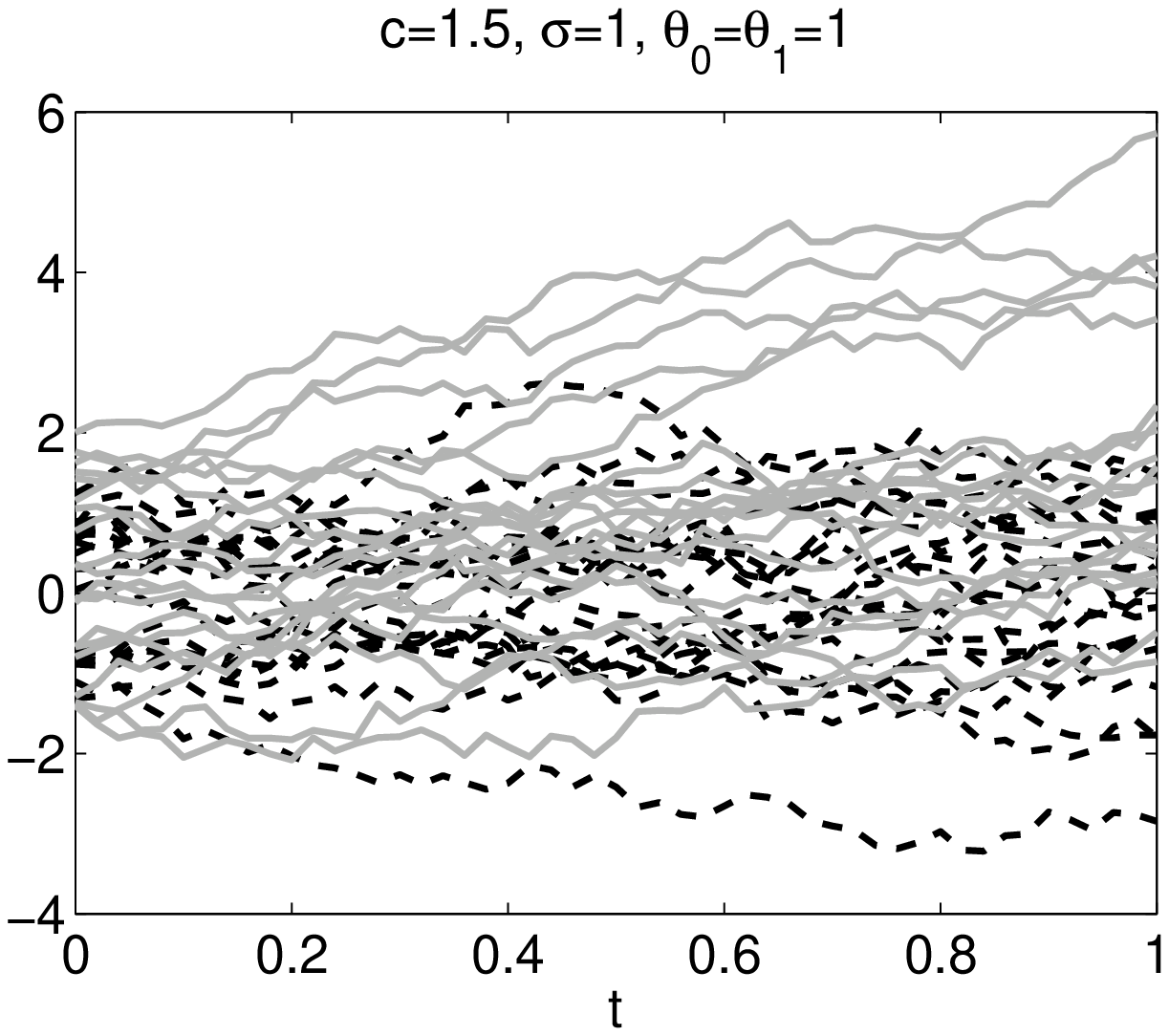}}
\\
(a) & (b) \\
\\
\resizebox{7 cm}{5 cm}{\includegraphics*[4cm,9.4cm][17cm,20.3cm]{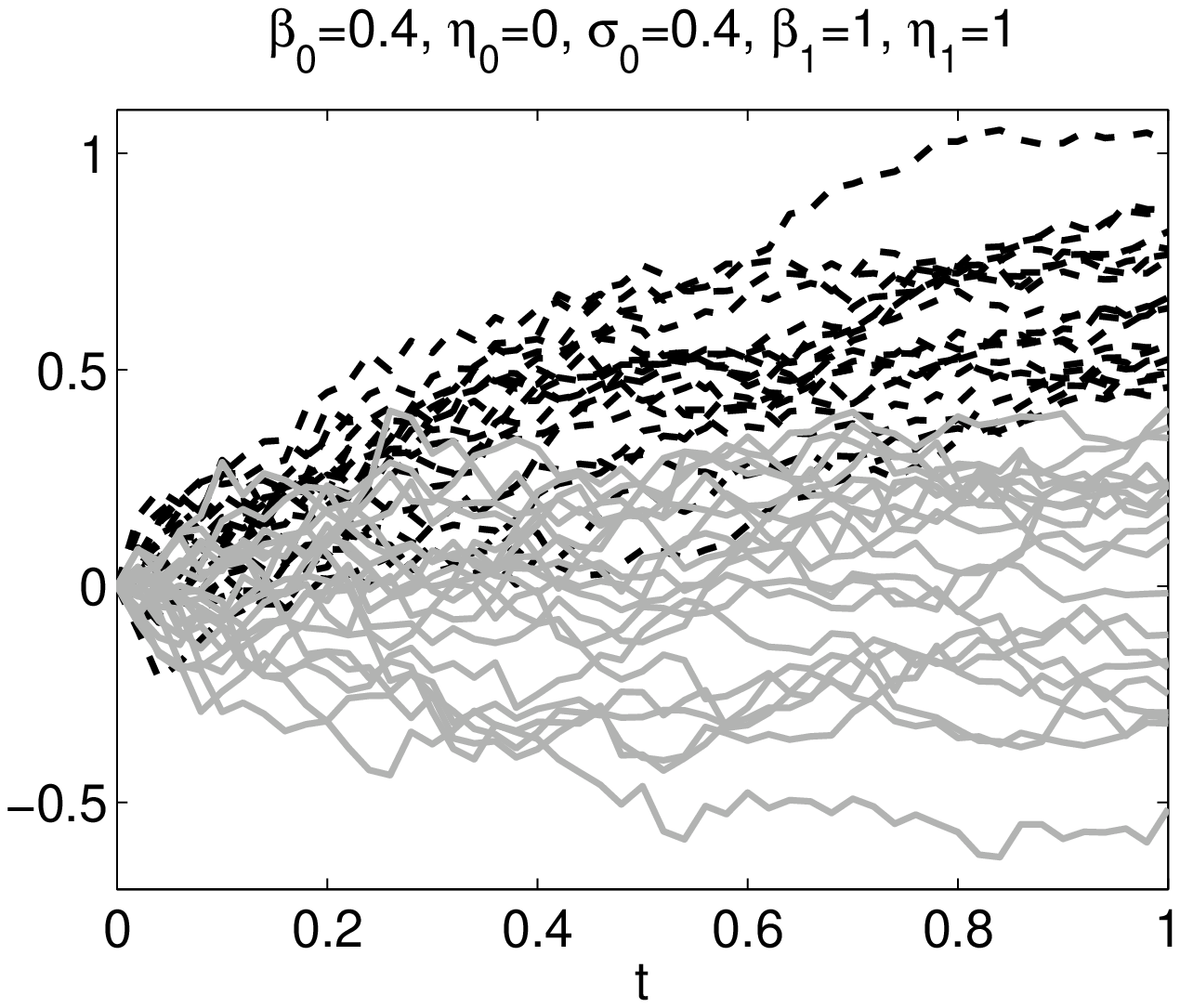}}
&
\resizebox{7 cm}{5 cm}{\includegraphics*[4cm,9.4cm][17cm,20.3cm]{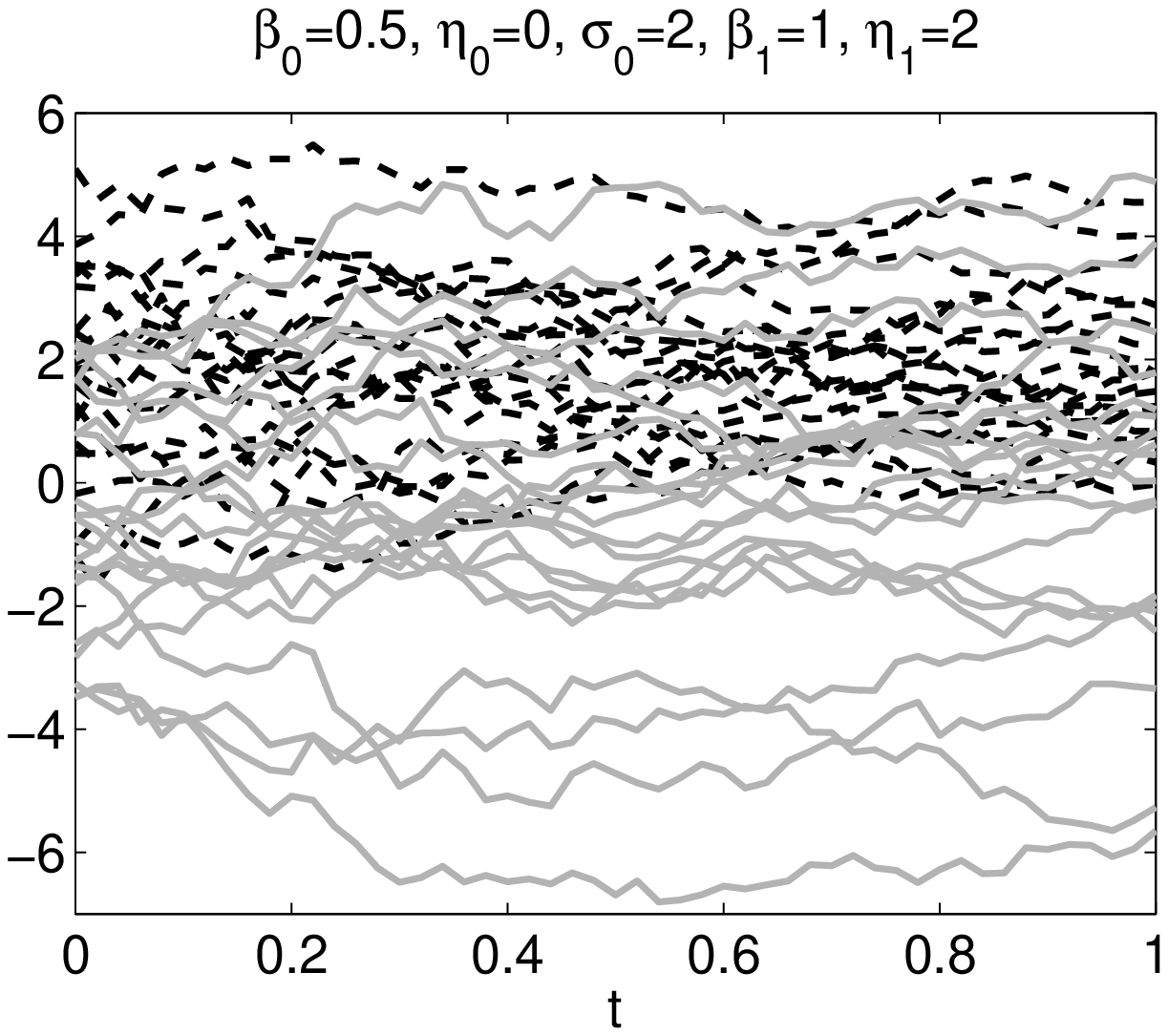}}
\\
(c) & (d)
\end{tabular}
\end{center}
\caption{Some trajectories ($P_0$ in gray and $P_1$ in dotted black) of the processes used in the Monte Carlo study.
In (a) and (b) we have two Brownian motions and in (c) and (d) the processes are Ornstein-Uhlenbeck. In (a) and (c) $X(0)|Y=i$ is 0 and in (b) and (d) it is random.}
\label{DibSimulations}
\end{figure}

The observations will be realizations of two Ornstein-Uhlenbeck processes and two Brownian motions
as described in Subsection~\ref{Subsection2_3}.
The parameters chosen for the pairs of processes
are specified in Table~\ref{table.simulation} (in Figure~\ref{DibSimulations}
we have depicted some trajectories of the processes used in the simulations).

\marginpar{\fbox{\hspace*{-2 mm}\begin{tabular}{l}
\tt Figure~\ref{DibSimulations} \\ \tt here.
\end{tabular}\hspace*{-3 mm}}}

We assume that $p={\mathbb P}\{ Y=0 \}$, the proportion of observations
coming from $P_0$, is $1/2$ and is known in advance.
For each $i=0,1$ we take a training sample
with size $n_i=100$ and a test sample with size $50$ from $P_i$.
The processes are observed at equidistant times of the interval $[0,1]$, $t_0=0, t_1,\ldots,t_N=1$, with $N=50$.
We denote by $\Delta=t_j-t_{j-1}$ the internodal distance.
The number of Monte Carlo runs is 1000.
In each run we use the training sample to construct four classifiers:
$k$-NN with the supremum norm and with a PLS-based semimetric (see e.g. Ferraty and Vieu, 2006, p. 30), 
parametric and nonparametric plug-in as introduced in Subsections \ref{Subsection2_3} and \ref{Subsection2_4b} respectively. 
The performance of these classifiers
is assessed by the proportion of correctly classified observations in the test samples.
We also compute this proportion for the Bayes rule associated to each model.
The number $k$ of neighbours and the number of PLS directions for projection are chosen
via cross-validation from a maximum of 10 neighbours and 5 PLS directions respectively.

When applying the nonparametric plug-in method to the data functions evaluated on the whole interval 
$[0,1]$ we observed a noticeable boundary effect near 0, especially in the estimation of $v$ and 
its derivatives. This made the nonparametric plug-in method perform poorly.
In order to avoid this, the Radon-Nikodym derivative for the nonparametric plug-in rule has been 
evaluated on the  trajectories restricted to the interval $[h_n,1]$, where $h_n$ is the same (and unique) 
smoothing parameter used in the estimation of the derivatives of $u_i$ and $v_i$. The value of $h_n$ 
has been chosen among $\{2\Delta,4\Delta,\ldots,20\Delta\}$ via cross-validation: for each 
$h_n=k\Delta$ we compute the corresponding estimated classification error 
with the usual leave-one-out device (every training observation is 
classified, as if it were a new incoming observation, using the remaining data as a training sample). 

In Table~\ref{table.simulation} we display the mean and the standard deviation (between parentheses)
of the proportion of correct classifications over the 1000 Monte Carlo samples.
We see that the parametric plug-in procedure is the one
performing best: it is very near the optimum.

As it could be expected, the nonparametric plug-in behaves worse than the parametric one. Its best performance corresponds to the random start cases $u_i(0)>0$ for $i=0,1$. In these situations, it is the second better classifier.  When $u_i(0)=0$, the parametric plug-in is still the winner,  the $k$-NN with PLS is the second and the  $k$-NN  with the supremum metric and the nonparametric plug-in  perform similarly.

It is interesting to note that the $k$-NN classification method is always reliable
(even with the supremum metric, although PLS semimetric yields better results).
Thus one of the conclusions of the study is that, when classifying functional data, the $k$-NN procedure is generally
a safe choice, free of model assumptions.

\marginpar{\fbox{\hspace*{-2 mm}\begin{tabular}{l}
\tt Table~\ref{table.simulation} \\ \tt here.
\end{tabular}\hspace*{-3 mm}}}

\subsection{A real data set} \label{Subsection4_2}
We compare the performance of the $k$-NN classification procedure with the
nonparametric plug-in one in the analysis of data
from research in experimental cardiology. The experiment was conducted at the Vall d'Hebron Hospital (Barcelona, Spain). See Ruiz-Meana {\em et al}.
(2003) for biochemical and medical details on the data and
Cuevas, Febrero and Fraiman (2004, 2006) for previous
analysis of these observations.

\begin{figure}
\begin{center}
\begin{tabular}{cc}
\resizebox{8 cm}{7 cm}{\includegraphics*[3.2cm,9.4cm][16.5cm,19.8cm]{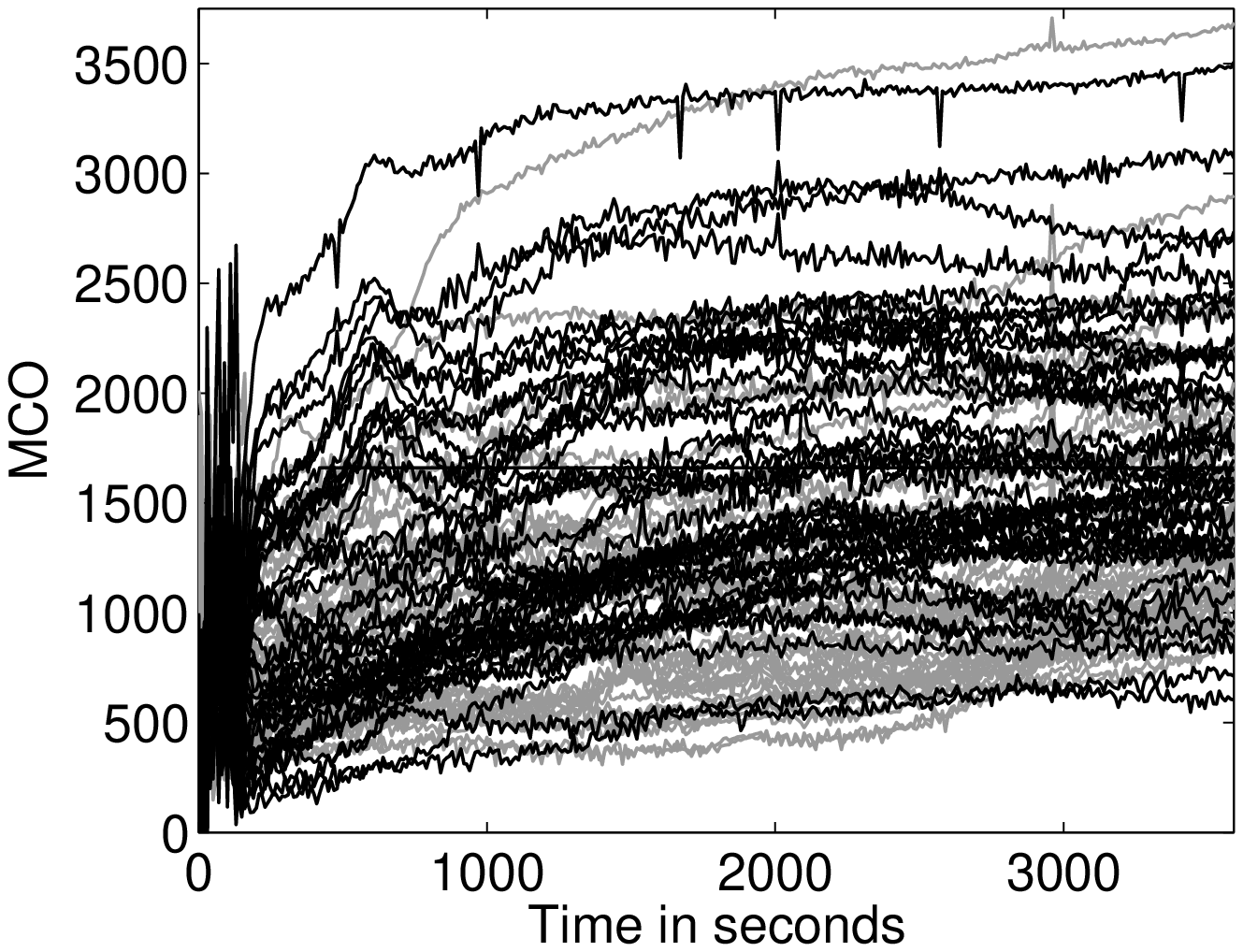}}
& \resizebox{8 cm}{7 cm}{\includegraphics*[3.2cm,9.4cm][16.5cm,19.8cm]{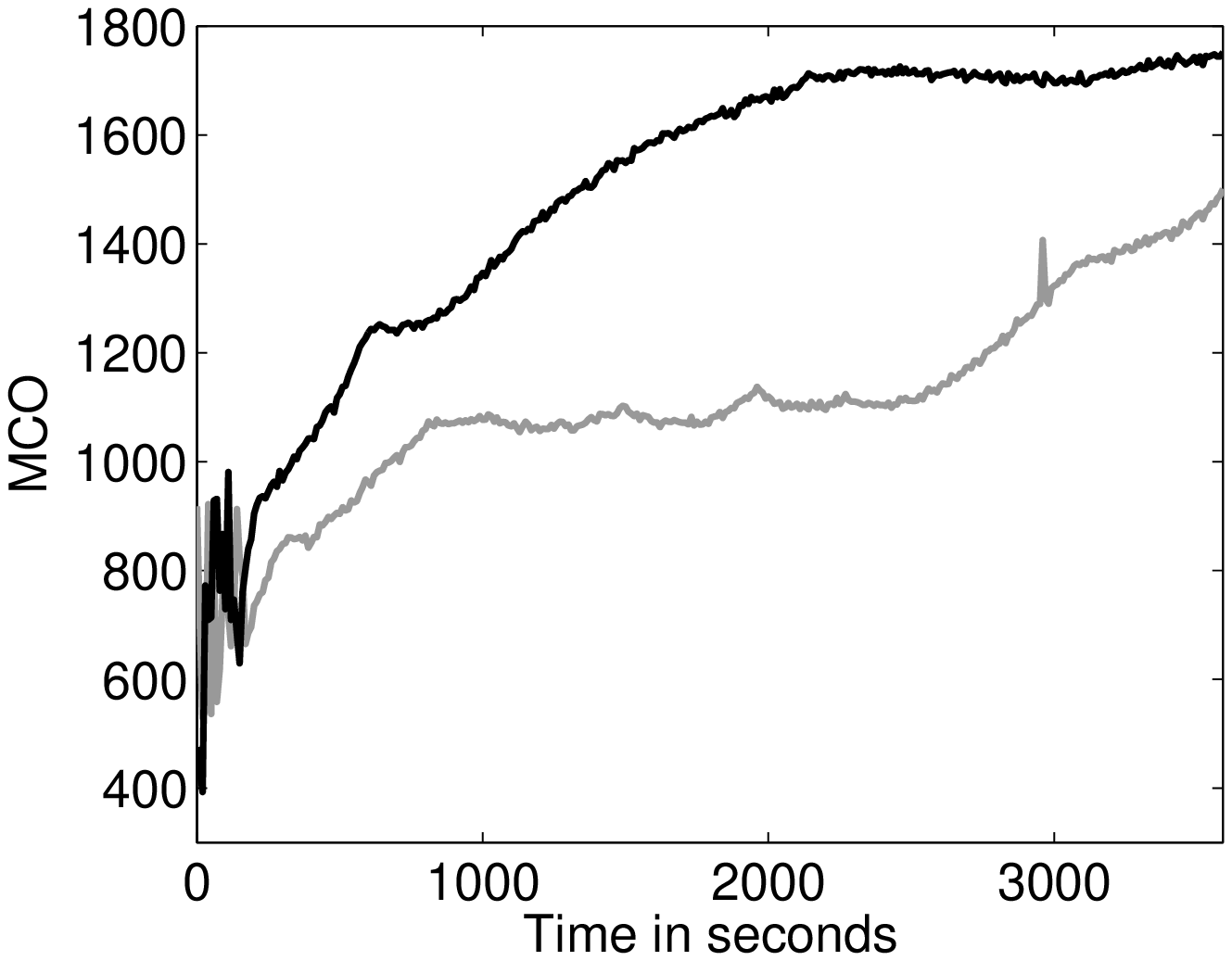}} \\
(a) & (b)
\end{tabular}
\end{center}
\caption{Cell data (control group in grey and treatment group in black): (a) all the original observations; (b) sample means.}
\label{CellsFigure1}
\end{figure}

The variable under study is the mitochondrial calcium overload (MCO), which measures the level of
the mitochondrial calcium ion (Ca2+). This variable was observed every 10 seconds
during an hour in isolated mouse cardiac cells.
The aim of the study was to assess whether a drug called Cariporide increased the MCO level.
The data we analyze here consist of two samples of functions with
sizes $n_0 = 45$ (control group) and $n_1 = 44$ (treatment group with Cariporide).
In Figure~\ref{CellsFigure1} we display (a) all the data
and (b) the group means.

\marginpar{\fbox{\hspace*{-2 mm}\begin{tabular}{l}
\tt Figure~\ref{CellsFigure1} \\ \tt here.
\end{tabular}\hspace*{-3 mm}}}

In many cases the first three minutes each curve shows
oscillations which correspond to normal contractions of the cells.
This first part of the curves has been eliminated (as in the original experiments with these data) because it has high
variability and depends on uncontrolled factors.

To obtain a better approach of the distributions to normality,
we have considered a transformation of the data, $X=\log(\mbox{MCO}-85)$.
The performance of any of the classification procedures considered is described by
the probability of correctly classifying one of the transformed observations, approximated
via cross-validation.

Obviously, in this case, we do not have enough information to consider using the parametric plug-in classifier.
Consequently we only employ  the $k$-NN (with uniform metric and PLS-based semimetric) and the nonparametric plug-in
discrimination rules. The results appear in Table~\ref{CellsTable2}. It is interesting to notice that the results in this case, in some sense, are the opposite to those obtained with the simulations.
The nonparametric plug-in  clearly outperforms the other two and the $k$-NN with the supremum metric does better than the $k$-NN with PLS.

\marginpar{\fbox{\hspace*{-2 mm}\begin{tabular}{l}
\tt Table~\ref{CellsTable2} \\ \tt here.
\end{tabular}\hspace*{-3 mm}}}

\vspace{3 mm}

\noindent
{\bf Acknowledgement.}
The authors want to thank Javier Segura for bringing to our knowledge some
numerical differentiation techniques (in particular, the $k$-point
rule).


\section{Appendix} \label{Appendix}

\noindent \bf A.1 Parameter estimation for the models of Subsection \ref{Subsection2_3}\rm

\noindent \it Two Brownian motions\rm

In the simulations of Section~\ref{Section4} the estimator of $c$ is
$\hat c={\arg\min}_c \sum_{j=1}^N (\hat m_0(t_j)-c\, t_j)^2$, where $m_i$ is the sample mean of
the observations coming from $P_i$.
The parameters $\theta_i$ and $\sigma^2$ are respectively estimated by $\hat\theta_i
= \sum_{j=1}^{n_i}\left( X_j(0;i)-\hat m_i(0) \right)^2/(n_i-1)$
and
$ \hat\sigma^2 = \sum_{i=0,1} \sum_{j=1}^{n_i}
\left( X_j(1;i)-\hat m_i(1)-X_i(0;i)+\hat m_i(0) \right)^2 /(n_0+n_1-1)$.
\vspace{2 mm}

\noindent
{\em Two Ornstein-Uhlenbeck processes}

The estimation of the unknown parameters ($\beta_i$, $\eta_i$ and $\sigma_i$, $i=0,1$) 
is carried out via linear least-squares regression between the realizations of the process at consecutive time points.
The main idea is that, for $i=0,1$ and for any $0\leq s<t\leq 1$, we have
\begin{equation} \label{OUupdate}
X(t;i) \, = \, X(s;i) \, e^{-\beta_i(t-s)} + \, \eta_i \, (1-e^{-\beta_i(t-s)})
+ \sigma_i\sqrt{1-e^{-2\beta_i(t-s)}} \, Z,
\end{equation}
where $Z$ is $N(0,1)$. The updating formula (\ref{OUupdate}) is valid when 
$X(0;i)$ is either deterministic or random. In particular, for $i=0,1$, $k=1,\ldots,n_i$ and $j=0,\ldots,N-1$,
\begin{equation} \label{LinearRegression}
X_{k}(t_{j+1};i) \, = \, a_i X_{k}(t_j;i) + b_i + \sigma_i\sqrt{1-e^{-2\beta_i\Delta}} \, Z_{kj},
\end{equation}
where $a_i:=e^{-\beta_i\Delta}$, $b_i:=\eta_i \, (1-e^{-\beta_i\Delta})$ and
$Z_{kj}$ are i.i.d. variables $N(0,1)$.

Observe that, by estimating the parameters of the simple linear regression equation
(\ref{LinearRegression}), we can construct estimators of $\beta_i$, $\eta_i$ and $\sigma_i$.
When $X(0;i)$ is deterministic, we compute the least-squares estimators of $a_i$ and $b_i$, that is, the values
$\hat a_i$ and $\hat b_i$ minimizing $ \sum_{k=1}^{n_i} \sum_{j=0}^{N-1} u_{kj}^2$,
where $ u_{kj} := X_{k}(t_{j+1};i) - (\hat a_i X_{k}(t_j;i) + \hat b_i)$ are the residuals.
Then
\begin{equation}
\hat\beta_i  =  -\frac{\log(\hat a_i)}{\Delta} , \qquad 
\hat\eta_i  =  \frac{\hat b_i}{1-\hat a_i} , \qquad
\hat\sigma_i^2  =  \frac{1}{(1-\hat a_i^2)(n_i N-2)} \sum_{k=1}^{n_i} \sum_{j=0}^{N-1} u_{kj}^2.
\label{OUEst}
\end{equation}

When $X(0;i)$ is random, we can compute $\hat\beta_i$ and $\hat\sigma_i^2$ as in (\ref{OUEst}),
but $\eta_i$ is better estimated by $\hat\eta_i=\sum_{j=1}^{n_i} \sum_{k=0}^{N} X_{ij}(t_k)/(n_i\, (N+1))$.

\

\noindent \bf A.2 Proofs of the results in \ref{Subsection2_4b}\rm

\noindent
{\sc Proof of Proposition \ref{TheoMedias}}

\noindent {\bf (a)} By the functional CLT in $(C [0,1],\| \cdot \|)$ (see p. 172 in Araujo and Gin\'e, 1980) 
the sequence $\sqrt{n}(\hat m_n-m)$ converges  weakly. This entails that the sequence $\|\sqrt{n}(\hat m_n-m)\|$ 
is bounded in probability  which in turn implies (\ref{TheoRes1}).
Concerning (\ref{TheoRes2}) and (\ref{TheoRes3}), let us denote
$X^*_i (t ) = X_i(t)- m(t), \ t \in [0,1], i=1,2,\ldots$.
 Note  that, for  $t \in [h_n,1-h_n]$,
\begin{eqnarray}
| m' (t)- \hat m_n' (t)|
&\leq&  \left|m' (t) -  \frac {m (t+h_n)- m (t-h_n)}{2h_n}\right|\nonumber
\\
& &
+ \left| \frac 1{2 h_n n} \sum_{i=1}^n X^*_i(t + h_n)\right|
+ \left| \frac 1{2 h_n n} \sum_{i=1}^n X^*_i(t - h_n)\right|\nonumber
\\
&\leq&  \left|m' (t) -  \frac {m (t+h_n)- m (t-h_n)}{2h_n}\right| 
+\left\| \frac 1{ h_n n} \sum_{i=1}^n X^*_i\right\|.\label{m-mn}
\end{eqnarray}
 The CLT applied to the sequence  $\{X^*_n \}$ allows us to conclude that the second term in the right-hand side of (\ref{m-mn}) is
$O_P\left((n^{1/2}h_n)^{-1}\right)$. A second order Taylor expansion of the first term implies that there exist $\psi_n^{(1)} \in (t-h_nt)$ and $\psi_n^{(2)}\in (t,t+h_n)$ such that
\[
\left|m' (t) -  \frac {m (t+h_n)- m (t-h_n)}{2h_n}\right| =
\frac{h_n}{4}
\left| m'' (\psi_n^{(1)}) - m''(\psi_n^{(2)}) \right|
\leq \frac{Lh_n^2}{4}=O(h_n^2),
\]
where $L$ is the Lipschitz constant associated with $m''$.

Applying a similar reasoning to (\ref{TheoRes3}), we obtain that, if $t \in [h_n,1-h_n]$, then,
\begin{equation}
| m'' (t)- \hat m_n'' (t)|
\leq \left|m'' (t) -  \frac {m (t+h_n) + m (t-h_n)- 2 m (t)}{h_n^2}\right|
+4 \left\| \frac 1{ h_n^2 n} \sum_{i=1}^n Y_i\right\|.\label{m2}
\end{equation}

The CLT implies that the order of the second term in (\ref{m2}) is $O_P\left((n^{1/2}h_n^2)^{-1}\right)$. A second order Taylor's expansion on $t$ again gives that
$$
  \left|m'' (t) -  \frac {m (t+h_n) + m (t-h_n) -2 m (t)}{h_n^2}\right|
   =
  \left|m'' (t) -  \frac 1 2 \left( m''(\psi_n^{(1)}) + m''(\psi_n^{(2)}) \right) \right|
     \leq  L h_n.
$$
\noindent {\bf (b)} Since
\begin{eqnarray*}
\hat \Gamma(t,1)  -  \Gamma(t,1)
& = &
\frac 1 n \left. \left. \sum_i \right( (X_i^*(t)+m(t)-\hat m_n (t))(X_i^*(1)+m(1)-\hat m_n (1))\right)-  \Gamma(t,1)
\\
& = &
\frac 1 n  \left. \left.  \sum_i \right( X_i^*(t)X_i^*(1)-  \Gamma(t,1) \right)
+
(m(t)-\hat m_n (t))\frac 1 n \sum_i X_i^*(1)
\\
&&
+(m(1)-\hat m_n (1)) \frac 1 n  \sum_i X_i^*(t)
+(m(t)-\hat m_n (t))(m(1)-\hat m_n (1)),
\end{eqnarray*}
then
\begin{eqnarray*}
\|\hat \Gamma(\cdot,1)  -  \Gamma(\cdot,1)  \|
& \leq &
\left\| \frac 1 n \sum_i \left( X_i^*X_i^*(1)-  \Gamma(\cdot,1) \right) \right\|
+
\left\| m-\hat m_n  \right\| \left| \frac 1 n \sum_i X_i^*(1)\right |
\\
&&
+ |m(1)-\hat m_n (1)| \left\| \frac 1 n \sum_i X_i^*\right\|
+\left\| m-\hat m_n  \right\| |m(1)-\hat m_n (1)|
\\
&=:&
 T_n^{(1)}+T_n^{(2)}+T_n^{(3)}+T_n^{(4)}.
\end{eqnarray*}
The assumption ${\mathbb E}\|X_1\|^4<\infty$ implies ${\mathbb E}\|X_i^*X_i^*(1)\|^2<\infty$ and thus 
the sequence $\{ X_i^*X_i^*(1)\}$ satisfies the CLT in the supremum norm. Then, since $E[ X_i^*X_i^*(1)]= \Gamma(\cdot,1) $,
we have that
$T_n^{(1)}=O_P(n^{-1/2})$.
Also $T_n^{(2)}=O_P(n^{-1})$ because the CLT (real case) implies that $ \sum_i X_i^*(1)/n=O_P(n^{-1/2})$  and, according to Proposition \ref{TheoMedias} (a), $\| m-\hat m_n\|=O_P(n^{-1/2})$.

The CLT applied to $\{X_i^*\}$ and Proposition \ref{TheoMedias} (a) yield that $T_n^{(3)}$ and 
$T_n^{(4)}$ are $O_P(n^{-1})$. This allows us to conclude (\ref{EqThV.u}).
The derivatives of $\Gamma(\cdot,1)$ are handled as those of $m$.
The estimators of $  \Gamma(0,\cdot)$ and $\sigma(\cdot)$ behave analogously to $  \Gamma(\cdot,1)$.
\hfill{$\Box$}

\

\noindent
{\sc Proof of Proposition \ref{TheoFuncv}}

\noindent{\bf a)} According to expression (\ref{Defn_n1}) for $\hat v_n(t)$, this estimator is a quotient of two convergent sequences. As that in the denominator, $\hat u_n(0)$,  converges to $u(0)>0$, an upper bound for the overall rate of the quotient is the slowest rate between $\hat\Gamma_n(0,t)$ and $\hat u_n(0)$. Similar arguments apply for the first and second derivatives.

\noindent {\bf b)} Let $t \in [\delta_n,1]$. The hypothesis on $u'$ implies that $\inf_{t \geq \delta_n} u(t) \geq O(\delta_n)$.  Since $n^{-1/2} = o(\delta_n)$, from (\ref{EqThV.u}) we obtain that $\inf_{t \geq \delta_n} \hat u_n (t) \geq O_P(\delta_n)$. Therefore, a direct calculation based on the expression of $\hat v_n$  together with Proposition \ref{TheoMedias} b) leads to
\begin{equation}\label{EqSigma0_0}
\sup_{t \in [\delta_n,1]} |\hat v_n (t) - v(t) |
= O_P\left(\frac{1}{\delta_n \sqrt n}\right).
\end{equation}
The same reasoning, taking into account the relative orders between $\delta_n$ and $h_n$ leads to
\begin{eqnarray}
\label{EqSigma0_1}
\sup_{t \in [\delta_n,1]} | \hat v_n' (t)- v' (t) | & = &
O_P\left(\frac{1}{\delta_n h_n \sqrt n}\right) + O\left(\frac{h_n^2}{\delta_n}\right)
\\
\label{EqSigma0_2}
\sup_{t \in [\delta_n,1]} | \hat v_n''(t)  - v'' (t) | & = & O_P\left(\frac{1}{\delta_n h_n^2 \sqrt n}\right) + O\left(\frac{h_n}{\delta_n^2}\right).
\end{eqnarray}

Now, let $t \in [0,\delta_n]$. Using the second-order Taylor expansion of $v$ at $\delta_n$, together with the definition (\ref{EqV_0_2}) of $\hat v_n$, we obtain that there exists $\psi_n \in (t,\delta_n)$ such that
\begin{eqnarray*}
\lefteqn{|\hat v_n (t) - v(t)|
\leq |\hat v_n(\delta_n) - v (\delta_n)| + (\delta_n-t) |\hat v_n'(\delta_n) - v' (\delta_n)|} \\
&&
\hspace*{5 mm} + \frac 1 2  (t-\delta_n)^2 |\hat v_n''(\delta_n)-v''(\delta_n)| + \frac 1 2  (t-\delta_n)^2 | v''(\delta_n)-v''(\psi_n)|
\\
& & \leq 
O_P\left(\frac{1}{\delta_n \sqrt n}\right)
+ O_P\left(\frac{1}{h_n \sqrt n}\right) + O\left({h_n^2}{}\right)
+  O_P\left(\frac{\delta_n}{ h_n^2 \sqrt n}\right) + O\left({h_n}{}\right)
+ O(\delta_n^3)
\\
& & = 
O_P\left(\frac{\delta_n}{ h_n^2 \sqrt n}\right) + O\left({h_n}{}\right)
+ O(\delta_n^3),
\end{eqnarray*}
where we have applied (\ref{EqSigma0_0}), (\ref{EqSigma0_1}) and (\ref{EqSigma0_2}) and the fact that $ v''$ is Lipschitz. Then the first statement in Proposition \ref{TheoFuncv} b) is deduced from here and (\ref{EqSigma0_0}).
The remaining two statements are proved similarly.
\hfill{$\Box$}
\vspace{3 mm}

Next we state a technical lemma which will be employed to prove Theorem \ref{TheoRatesC}.

\begin{Lemm} \label{LemmTheoRates}
Let $\{Y(t), t\in [0,1]\}$ be a stochastic process whose mean function $m(t)$ and variance function  $\sigma^2 (t) $ satisfy that  $m(0)=\sigma(0)=0$ and both have  a bounded derivative. Let $\{\delta_n\}$ be positive numbers which converge to zero. Then
\[
{\mathbb E}\int_0^{\delta_n} |Y(t)| dt = O(\delta_n^{3/2})\quad \mbox{ and } \quad{\mathbb E}\int_0^{\delta_n} Y^2(t) dt = O(\delta_n^{2}) .
\]
\end{Lemm}
{\sc Proof:} Let $H$ be a common upper bound for the derivatives of $m^2$ and $\sigma^2$.
\begin{eqnarray*}
\int_0^{\delta_n} {\mathbb E}|Y(t)| dt
&\leq&
  \int_0^{\delta_n} {\mathbb E}^{1/2}(Y^2(t)) dt
= \int_0^{\delta_n} (m(t)^2 + \sigma^2(t))^{1/2} dt\\
 &   \leq &
  (2H)^{1/2} \int_0^{\delta_n} t^{1/2} dt =  O(\delta_n^{3/2}).
  \end{eqnarray*}
The second statement in the lemma follows analogously.
\hfill{$\Box$}
\vspace{3 mm}

\noindent
{\sc Proof of Theorem \ref{TheoRatesC}:}
From expressions (\ref{partea}) and (\ref{parteb}) we see that $f=d\mu_0/d\mu_1$ is a function of $m_i$, $u_i$, $v_i$ and their derivatives. Statement a) corresponds to the simplest case in which $u_i(0)>0$. In this situation, the simple structure of the estimators shows that an upper bound for the convergence rate for $\log f_n(x)$ is the worst rate for the estimators involved in its definition, namely that of the estimators $v_0''$ and $v_1''$.

Hence, we concentrate on part b). For simplicity we will omit the sub-index in $v_i$ for the rest of the proof. First notice that  in the expressions for ${d\mu_0}/{d \mu_1}$ which we obtained in
Theorem \ref{TheoExpresion} the second derivatives of $v$ only appear inside integrals. In other words, we only need to care about differences of the type
\begin{equation}
\int_0^1 X^r(t) ( \hat k_n (t) \hat v''_n (t) - k (t) v'' (t) ) \, dt
=
O_P\left( \int_0^1 X^r(t) k (t) (\hat v''_n (t) - v'' (t)) dt  \right),\label{split}
\end{equation}
for $r=1,2$. Here $k$ is a function depending on $u, v, u', v', m$ and $m'$ and $X$ is a mixture of the Brownian motions under consideration. Let us analyze the case in Theorem \ref{TheoExpresion} b) for which $r=1$ and the function $k$ can be expressed as
$k= k_1/\left(v((vu'-uv')^2\right)$,
where $k_1$ is a function which can be written in terms of $u, v, u', v', m$ and $m'$. Therefore, the assumptions  in Theorem \ref{TheoExpresion}, imply that $k$ is bounded.  Let $K$ be an upper bound of $k$.

We split in two the integral in the right-hand side of (\ref{split}), over the intervals $[0,\delta_n]$ and $[\delta_n,1]$. Now, from (\ref{EqSigma0_2}) in the proof of Proposition \ref{TheoFuncv}, we have that
\begin{eqnarray} 
\lefteqn{{\mathbb E}\left(\left.| \int_{\delta_n}^1 X(t) k (t)( \hat v''_n (t) - v'' (t) ) dt| \right|X_1,\ldots,X_n\right)}\nonumber\\
&&\hspace*{1 cm}\leq \left(O_P\left(\frac{1}{\delta_n h_n^2 \sqrt n}\right) + O\left(\frac{h_n}{\delta_n^2}\right)\right)
\left(\int_{\delta_n}^1 {\mathbb E}(X^2(t))dt\right)^{1/2}.\label{EqTheRates1}
\end{eqnarray}

With respect to the other integral, we have that
\begin{eqnarray}
\nonumber
\lefteqn{{\mathbb E}\left(\left.| \int_0^{\delta_n} X(t) k (t) (\hat v''_n (t) - v'' (t) ) dt|\right|X_1,\ldots,X_n\right)}\\
&&\leq 
K\,\| \hat v''_n  - v''  \|\ {\mathbb E} \int_0^{\delta_n} |X(t)|  dt
\label{EqTheRates2}
=
O_P\left(\frac{\delta_n^{1/2}}{ \sqrt n}\right) + O\left(\frac{h_n}{\delta_n^{1/2}}\right) + O(\delta_n^{5/2}),
\end{eqnarray}
where the last equality comes from Lemma \ref{LemmTheoRates} and Proposition \ref{TheoFuncv} b).
Equations (\ref{EqTheRates1}) and (\ref{EqTheRates2}) give 
$$
{\mathbb E}\left(\left.| \int_{0}^1 X(t) k (t)( \hat v''_n (t) - v'' (t) ) dt| \right|X_1,\ldots,X_n\right)
\leq 
O_P\left(\frac{1}{\delta_n h_n^2 \sqrt n}\right) + O\left(\frac{h_n}{\delta_n^2}\right)
+
O(\delta_n^{5/2}).
$$
Taking $h_n = \delta_n^{9/2}$ and $\delta_n = n^{-1/25}$ equates the three terms and yields the result. 
\hfill{$\Box$}

\

\noindent
{\sc Proof of Proposition
\ref{CoroTheoRatesC}:}
It follows the same steps as the proof of Proposition \ref{TheoMedias}, the only difference being that if we apply a third order Taylor expansion in (\ref{m2}), we obtain
\begin{eqnarray*}
  \left|m'' (t) -  \frac {m (t+h_n) + m (t-h_n) -2 m (t)}{h_n^2}\right|
  & = &
 \frac {h_n} {3!} \left| \left( m'''(\psi_n^1) - m'''(\psi_n^2) \right) \right|
     \leq  \frac{L h_n^2}{3!},
\end{eqnarray*}
and the result follows.
\hfill{$\Box$}

\

\noindent
{\sc Proof of Theorem \ref{latasaparaelerror}:}
Let us use the following inequality (see, e.g., Devroye {\em et al}., 1996, p. 93)
$$
\hat L_n - L^* \leq 2 \; \mathbb E \left( \left. |\eta(X)-\eta_n(X)| \ \right| \mathcal X_n \right) ,
$$
where $\eta$ is given in (\ref{EqLareglaoptima}) and $\eta_n$ is obtained substituting $f=d\mu_0/d\mu_1$ by $\hat f_n$ in (\ref{EqLareglaoptima}).
Without loss of generality in this proof we consider $p=\mathbb P\{ Y=0 \} =1/2$.

Observe that, $f$ and $\hat f_n$ are always positive since they are Radon-Nikodym derivatives of one probability measure with respect to
another.
Thus, for any $x$, we have
$$
|\eta(x)-\eta_n(x)| = \frac{|f(x)-\hat f_n(x)|}{(1+\hat f_n(x)(1+f(x))} \leq |f(x)-\hat f_n(x)|,
$$
which implies that
\begin{equation} \label{Ineq1}
\hat L_n - L^* \leq 2 \; \mathbb E \left( \left. |f(x)-\hat f_n(x)| \right| \mathcal X_n \right).
\end{equation}
We obtain convergence rates (in probability) for the conditional expectation in the right of (\ref{Ineq1}).
Since all the cases are similar,  let us consider the simple situation in which $m_0\neq m_1$ and $\Gamma_0=\Gamma_1=\Gamma$ with
$\Gamma(s,t)=u(\min(s,t)) \, v(\max(s,t))$. Then
\begin{equation} \label{DifeDeriv}
f-\hat f_n = \frac{dP_{m_0,\Gamma}}{dP_{m_1,\Gamma}} -  \frac{dP_{\hat m_0,\hat\Gamma_0}}{dP_{\hat m_1,\hat\Gamma_1}}
= \frac{dP_{m_0,\Gamma}}{dP_{m_1,\Gamma}} -  \frac{dP_{\hat m_0,\hat\Gamma_0}}{dP_{\hat m_1,\hat\Gamma_0}}
+ \frac{dP_{\hat m_0,\hat\Gamma_0}}{dP_{\hat m_1,\hat\Gamma_0}}
\left( 1- \frac{dP_{\hat m_1,\hat\Gamma_0}}{dP_{\hat m_1,\hat\Gamma_1}} \right).
\end{equation}

By Theorem~\ref{TheoExpresion} (b) and the mean value theorem we have that, for any $x$,
$$
\frac{dP_{m_0,\Gamma}}{dP_{m_1,\Gamma}}(x) -  \frac{dP_{\hat m_0,\hat\Gamma_0}}{dP_{\hat m_1,\hat\Gamma_0}}(x)
= e^{z} (z_1-z_2),
$$
where (using the notation of Theorem~\ref{TheoExpresion})
\begin{eqnarray*}
z_1
&=&
 D_1  + \left( D_2 -2\,\frac{G(0)}{v(0)} \right) x(0) + 2\,\frac{G(1)}{v(1)} \, x(1) -2 \int_0^1 \frac{x(t)}{v(t)} \, G'(t) \, dt ,
\\
z_2
 &=&
 \hat D_{1;0} + \left( \hat D_{2;0} -2\,\frac{\hat G(0)}{\hat v_0(0)} \right) x(0)
 + 2 \, \frac{\hat G_0(1)}{\hat v_0(1)} \, x(1)  -2 \int_0^1 \frac{x(t)}{\hat v_0(t)} \, \hat G_0'(t) \, dt
\end{eqnarray*}
and $z=\lambda \, z_1+(1-\lambda)z_2$ for some $\lambda\in[0,1]$.
The subscripts 0 in the expression of $z_2$ mean that the estimation is carried out only with the sample from $P_0$.

Consequently,
\begin{eqnarray}
\lefteqn{\mathbb E \left( \left.| \frac{dP_{m_0,\Gamma}}{dP_{m_1,\Gamma}}(X)
- \frac{dP_{\hat m_0,\hat\Gamma_0}}{dP_{\hat m_1,\hat\Gamma_0}} (X)  | \right|\mathcal X_n \right)  } \nonumber \\
 & & \leq \mathbb E \left\{ e^{|Z_1|+|Z_2|} \left[ |D_1-\hat D_{1;0}|
     + \left( |D_2 - \hat D_{2;0}| + 2 \left| \frac{G(0)}{v(0)}
     - \frac{\hat G(0)}{\hat v_0(0)} \right| \right) |X(0)| \right. \right. \nonumber \\
 & & \hspace{5 mm} \left. \left. + 2 \left| \frac{G(1)}{v(1)} - \frac{\hat G_0(1)}{\hat v_0(1)} \right| |X(1)|
 + 2\int_0^1 |X(t)| \left.\left| \frac{G'(t)}{v(t)} - \frac{\hat G_0'(t)}{\hat v_0(t)} \right| dt \right] \right| \mathcal X_n \right\}
 \label{CondExpec1} \\
 & & \leq \kappa \left\{ |D_1-\hat D_{1;0}| \, \mathbb E \left( e^{A\|X\|} \, | \mathcal X_n \right)
 + \left( |D_2 - \hat D_{2;0}|
     + 2 \max_{t=0,1} \left| \frac{G(t)}{v(t)} - \frac{\hat G_0(t)}{\hat v_0(t)} \right|
 \right. \right. \label{CondExpec2} \\
 & & \hspace{5 mm} \left.   \left.
     + 2 \int_0^1 \left| \frac{G'(t)}{v(t)} - \frac{\hat G_0'(t)}{\hat v_0(t)} \right| dt \right)
     \mathbb E \left( \|X\| e^{A\|X\|} | \mathcal X_n \right) \right\}  \label{CondExpec5}
\end{eqnarray}
where $\kappa = \exp(|D_1|+|\hat D_{1;0}|)$ and
$$
A = \max \left( |D_2|+|\hat D_{2;0}| , \left\| \frac{G}{v} + \frac{\hat G_0}{\hat v_0} \right\|,
\left\| \frac{G'}{v} + \frac{\hat G_0'}{\hat v_0} \right\| \right).
$$
Using Propositions~\ref{TheoMedias} and~\ref{TheoFuncv} we obtain that the conditional expectations appearing in
(\ref{CondExpec2}) and (\ref{CondExpec5}) are bounded in probability. Then
\begin{eqnarray*}
\lefteqn{\mathbb E \left( \left. | \frac{dP_{m_0,\Gamma}}{dP_{m_1,\Gamma}}(X)
         - \frac{dP_{\hat m_0,\hat\Gamma_0}}{dP_{\hat m_1,\hat\Gamma_0}} (X)  |\right| \mathcal X_n \right)
         = O_P\left(\max_{j=1,2} |D_j-\hat D_{j;0}|\right) } \\
 & & 
 + \, O_P\left(\max_{t=0,1} \left| \frac{G(t)}{v(t)} - \frac{\hat G_0(t)}{\hat v_0(t)} \right|\right)
     + O_P \left(\int_0^1 \left| \frac{G'(t)}{v'(t)} - \frac{\hat G_0'(t)}{\hat v_0'(t)} \right| dt \right).
\end{eqnarray*}
To find the convergence rates to 0 of these last three terms  we use the expressions of $D_1$, $D_2$ and $G$ appearing in Theorem~\ref{TheoExpresion}.
Some straighforward computations yield
$|D_1-\hat D_{1;0}|=O_P(\| \hat v_0'-v' \|)$, $|D_2-\hat D_{2;0}|=O_P(\| \hat v_0-v \|)$,
$$
\max_{t=0,1} \left| \frac{G(t)}{v(t)} - \frac{\hat G_0(t)}{\hat v_0(t)} \right| = O_P(\| \hat v_0'-v' \|)
\quad
\mbox{and}
\quad
\int_0^1 \left| \frac{G'(t)}{v'(t)} - \frac{\hat G_0'(t)}{\hat v_0'(t)} \right| dt
= O_P(\| \hat v_0''-v'' \|).
$$
Thus we get
\begin{equation} \label{Parte1}
\mathbb E \left( \left.| \frac{dP_{m_0,\Gamma}}{dP_{m_1,\Gamma}}(X)
- \frac{dP_{\hat m_0,\hat\Gamma_0}}{dP_{\hat m_1,\hat\Gamma_0}} (X)  |\right| \mathcal X_n \right)
= O_P(\| \hat v_0''-v'' \|).
\end{equation}

Let us now focus on the last term of (\ref{DifeDeriv}). The analysis is similar to the one carried out above. On the one hand,
for any $x$ it holds that
$$
\frac{dP_{\hat m_0,\hat\Gamma_0}}{dP_{\hat m_1,\hat\Gamma_0}} (x) \leq \kappa \, e^{2B\|x\|},
$$
where $B = \max ( |\hat D_{2;0}| , \| \hat G_0/\hat v_0 \|, \| \hat G_0'/\hat v_0 \| )$.
On the other hand, for any $x$ it also holds that
\begin{eqnarray}
\lefteqn{ \left| 1- \frac{dP_{\hat m_1,\hat\Gamma_0}}{dP_{\hat m_1,\hat\Gamma_1}} (x) \right| } \nonumber \\
 & & \leq |C_1-\hat C_1| + \frac{1}{2} \, \hat C_1 \, e^{\Lambda \|x\|^2}
 \left( |\hat C_3|x^2(0) + |\hat C_2|x^2(1) + \int_0^1 x^2(t) \frac{|\hat F'(t)|}{\hat v_0(t) \hat v_1(t)} \, dt \right)
 \label{Integral2} \\
 & & \leq |C_1-\hat C_1| + \hat C_1 \, \Lambda \, e^{\Lambda \|x\|^2} \|x\|^2, \nonumber
\end{eqnarray}
where 
$ \Lambda = ( |\hat C_3| + |\hat C_2| +\int_0^1 |\hat F'|/(\hat v_0 \hat v_1) )/2 $.
Consequently
\begin{eqnarray}
\lefteqn{ \mathbb E \left( \frac{dP_{\hat m_0,\hat\Gamma_0}}{dP_{\hat m_1,\hat\Gamma_0}} (X)
\left.| 1- \frac{dP_{\hat m_1,\hat\Gamma_0}}{dP_{\hat m_1,\hat\Gamma_1}} (X) | \right| \mathcal X_n \right)
} \nonumber  \\
 & & \leq \kappa \left\{ |C_1-\hat C_1| \; \mathbb E\left( e^{2B\|X\|} | \mathcal X_n  \right)
 + \, \hat C_1 \, \Lambda \, \mathbb E\left(\left. \|X\|^2 \,
 e^{2B\|X\|+\Lambda\|X\|^2} \right| \mathcal X_n \right) \right\} . \label{CondExpec8}
\end{eqnarray}
The conditional expectations in (\ref{CondExpec8}) and $\hat C_1$ are $O_P(1)$.
The term $\Lambda$ is $O_P(\max_{j=0,1} \|\hat v_j''-v''\|)$. The difference $|C_1-\hat C_1|$ is
$O_P(\max_{j=0,1} \|\hat v_j-v\|)$. Thus the term in (\ref{CondExpec8}) is
$O_P(\max_{j=0,1} \|\hat v_j''-v''\|)$. This, together with (\ref{Parte1}) and Proposition~\ref{TheoFuncv} (a), yield the desired result.
\hfill $\Box$

\


\noindent
{\bf \large References}

\begin{list}{}{\itemsep=-1mm \leftmargin=5 mm \rightmargin=0 mm \itemindent=-5 mm}
\item Abraham, C., Biau, G. and Cadre, B. (2006). On the kernel rule for function classification.
{\em Ann. Inst. Stat. Math.} {\bf 58}, 619--633.
\item Araujo A. and Gin\'e, E.  (1980).
{\em The Central Limit Theorem for Real and Banach Valued Random Variables.} Wiley.
\item Audibert, J.Y. and Tsybakov, A.B. (2007). Fast learning rates for plug-in classifiers. {\em Ann. Statist.} \bf 35\rm, 608--633.
\item  Ba\'{\i}llo, A\rm.,  Cuevas, A\rm. and Fraiman, R\rm. (2009). Classification methods for functional data. 
To appear in \it Oxford Handbook on Statistics and FDA\rm, F. Ferraty and Y. Romain, eds. Oxford University Press.
\item C\'erou, F. and Guyader, A. (2006). Nearest neighbor classification in infinite dimension.
{\em ESAIM Probab. Stat.} \bf 10\rm, 340-355.
\item Cuevas, A., Febrero, M. and Fraiman, R. (2004). An anova test for functional data. {\em Comput. Statist. Data Anal.}  \bf 47\rm, 111--122.
\item Cuevas, A., Febrero, M. and Fraiman, R. (2006). On the use of the bootstrap for estimating functions 
with functional data. {\em Comput. Statist. Data Anal.}  \bf 51\rm, 1063--1074.
\item Devroye, L., Gy{\"o}rfi, L. and Lugosi, G. (1996). {\em A Probabilistic Theory of
Pattern Recognition}. Springer, New York.
\item Duda, R.O., Hart, P.E., Stork, D.G. (2000). {\em Pattern Classification, 2nd edition}. Wiley.
\item Ferraty, F. and Vieu, P. (2006). {\em Nonparametric Functional Data Analysis: Theory and Practice}. Springer.
\item Feldman, J. (1958). Equivalence and perpendicularity of Gaussian processes. {\em  Pacific J. Math.} \bf 8\rm, 699--708.
\item Fisher, R.A. (1936). The use of multiple measurements in taxonomic problems.
{\em Annals of Eugenics} {\bf 7}, 179--188.
\item Folland, G. B. (1999). {\em Real Analysis Modern Techniques and their Applications}.
Wiley, New York.
\item Gautschi, W. (1997). {\em Numerical Analysis. An Introduction}. Birkh\"auser. Boston.
\item Hand, D. (2006). Classifier technology and the illusion of progress. \it Statist. Sci. \bf 21\rm, 1--34. 
\item Hastie, T., Tibshirani, R. and Friedman, J. (2001). {\em The Elements of Statistical Learning.\/}
Springer. New York.
\item Herzeg, D. and Cvetkovic, L. (1986). On a numerical differentiation. {\em SIAM J. Numer. Anal.} 
{\bf 23}, 686--691.
\item James, G. M., Hastie, T. J. (2001). Functional linear discriminant analysis
for irregularly sampled curves. {\em J. Roy. Statist. Soc. Ser. B}
{\bf 63}, 533--550.
\item J\o rsboe, O. G. (1968). {\em Equivalence or Singularity of Gaussian Measures on
Function Spaces}. Various Publications Series, No. 4, Matematisk Institut, Aarhus
Universitet, Aarhus.
\item Ramsay, J.O. and Silverman, B.W. (2005). {\em Functional Data Analysis}. Second edition. Springer.
\item Ruiz-Meana, M.,  Garc\'{\i}a-Dorado, D., Pina, P.,
Inserte, J., Agull\'o, L. and Soler-Soler, J.  (2003).
Cariporide preserves mitochondrial proton gradient and
delays ATP depletion in cardiomyocites during ischemic
conditions. \it Am. J. Physiol. Heart Circ. Physiol.  \bf 285\rm, H999-H1006.

\item Sacks, J. and Ylvisaker, N.D. (1966). Designs for regression problems with correlated errors.
{\em Ann. Math. Statist.} \bf 37\rm, 66--89.
\item Segall, A. and Kailath, T. (1975). Radon-Nikodym derivatives with respect to measures induced by discontinuous
independent-increment processes. {\em Ann. Probab.} \bf 3\rm, 449--464.

\item
Shin, J. (2008). An extension of Fisher's discriminant analysis for
stochastic processes. {\em J. Multiv. Anal.} \bf 99\rm, 1191--1216.

\item Stone, C. J. (1977). Consistent nonparametric regression. {\em Ann. Statist.} \bf 5\rm, 595--645.
\item Vakhania, N.N. (1975). The topological support of Gaussian measure in Banach space.
{\em Nagoya Math. J.} \bf 57\rm, 59--63.
\item Varberg, D.E. (1961). On equivalence of Gaussian measures.
{\em  Pacific J. Math.} \bf 11\rm, 751--762.
\item Varberg, D.E. (1964). On Gaussian measures equivalent to Wiener measure. {\em Trans.
Amer. Math. Soc.} \bf 113\rm, 262--273.

\end{list}

\newpage

\begin{table}
\begin{center}
\begin{tabular}{|@{\hspace{0mm}}c@{\hspace{0mm}}|@{\hspace{0mm}}c@{\hspace{0mm}}|c@{\hspace{0mm}}|c@{\hspace{3mm}} c @{\hspace{-1mm}} c @{\hspace{-2.5mm}} c @{\hspace{-1mm}} c@{\hspace{-1mm}}|} \cline{4-8}
\multicolumn{3}{c|}{} & \begin{tabular}{@{\hspace{0mm}}c@{\hspace{0mm}}} $k$-NN \\ $\|\;\|_\infty$ \end{tabular}
 & \begin{tabular}{c} $k$-NN \\ PLS \end{tabular}
 & \begin{tabular}{c} Nonpar. \\ plug-in \end{tabular}
 & \begin{tabular}{c} Param. \\ plug-in \end{tabular}
 & \begin{tabular}{c} Bayes \\ rule \end{tabular} \\ \hline
\multirow{10}{*}{\begin{tabular}{@{\hspace{0mm}}c@{\hspace{0mm}}} Two \\ Brownian \\ motions \end{tabular}}
 & \multirow{6}{*}{\begin{tabular}{@{\hspace{0.5mm}}c@{\hspace{0.5mm}}} Deterministic \\ at $t=0$ \\ ($\theta_0=\theta_1=0$) \end{tabular}}
     & \multirow{2}{*}{\begin{tabular}{c} $c=1.5$, $\sigma=1$ \end{tabular}} & 0.68 & 0.73 & 0.71 & 0.77 & 0.77 \\
     & & & (0.07) & (0.07) & (0.16) & (0.06) & (0.06) \\ \cline{3-8}
     & & \multirow{2}{*}{\begin{tabular}{c} $c=3$, $\sigma=1$ \end{tabular}} & 0.90 & 0.91 & 0.86 & 0.93 & 0.93 \\
     & & & (0.05) & (0.05) & (0.16) & (0.04) & (0.03) \\ \cline{3-8}
     & & \multirow{2}{*}{\begin{tabular}{c} $c=2$, $\sigma=2$ \end{tabular}} & 0.60 & 0.64 & 0.64 & 0.69 & 0.69 \\
     & & & (0.08) & (0.08) & (0.16) & (0.07) & (0.06) \\ \cline{2-8}
 & \multirow{4}{*}{\begin{tabular}{c} Random \\ at $t=0$ \\ ($\theta_0,\theta_1\neq 0$) \end{tabular}}
 & \multirow{2}{*}{\begin{tabular}{c} $c=1.5$, $\sigma=1$ \\ $\theta_0=\theta_1=1$ \end{tabular}} & 0.67 & 0.66 & 0.71 & 0.77 & 0.77 \\
 & & & (0.07) & (0.08) & (0.08) & (0.07) & (0.06) \\ \cline{3-8}
 & & \multirow{2}{*}{\begin{tabular}{c} $c=1.5$, $\sigma=1$ \\ $\theta_0=\theta_1=0.5$ \end{tabular}} & 0.67 & 0.70 & 0.72 & 0.77 & 0.77 \\
 & & & (0.07) & (0.08) & (0.08) & (0.06) & (0.06) \\ \hline
 \multirow{8}{*}{\begin{tabular}{@{\hspace{0.5mm}}c@{\hspace{0.5mm}}} Two \\ Ornstein- \\ Uhlenbeck \\ processes \end{tabular}}
 & \multirow{4}{*}{\begin{tabular}{@{\hspace{0.5mm}}c@{\hspace{0.5mm}}} Deterministic \\ at $t=0$ \end{tabular}}
 & \multirow{2}{*}{\begin{tabular}{@{\hspace{0mm}}c@{\hspace{0mm}}} $\beta_0 = 1$, $\eta_0 = 0$, $\sigma_0 = 1$ \\ $\beta_1 = 1$, $\eta_1 = 1$ \end{tabular}}
 & 0.54 & 0.58 & 0.60 & 0.63 & 0.62 \\
 & & & (0.08) & (0.08) & (0.14) & (0.07) & (0.07) \\ \cline{3-8}
 & & \multirow{2}{*}{\begin{tabular}{@{\hspace{0mm}}c@{\hspace{01mm}}} $\beta_0 = 0.4$, $\eta_0 = 0$, $\sigma_0 = 0.4$ \\ $\beta_1 = 1$, $\eta_1 = 1$ \end{tabular}}
 & 0.83 & 0.86 & 0.82 & 0.88 & 0.88 \\
 & & & (0.09) & (0.06) & (0.16) & (0.05) & (0.05) \\ \cline{2-8}
 & \multirow{4}{*}{\begin{tabular}{c} Random \\ at $t=0$ \end{tabular}}
 & \multirow{2}{*}{\begin{tabular}{@{\hspace{0mm}}c@{\hspace{0mm}}} $\beta_0 = 0.5$, $\eta_0 = 0$, $\sigma_0 = 1$ \\ $\beta_1 = 1$, $\eta_1 = 0.5$ \end{tabular}}
 & 0.59 & 0.60 & 0.63 & 0.63 & 0.64 \\
 & & & (0.13) & (0.11) & (0.14) & (0.07) & (0.14) \\ \cline{3-8}
 & & \multirow{2}{*}{\begin{tabular}{@{\hspace{0mm}}c@{\hspace{0mm}}} $\beta_0 = 0.5$, $\eta_0 = 0$, $\sigma_0 = 2$ \\ $\beta_1 = 1$, $\eta_1 = 2$ \end{tabular}}
 & 0.69 & 0.72 & 0.74 & 0.74 & 0.74 \\
 & & & (0.11) & (0.10) & (0.11) & (0.07) & (0.09) \\ \hline
\end{tabular}
\end{center}
\caption{Results of the Monte Carlo study} \label{table.simulation}
\end{table}

\begin{table}
\begin{center}
\begin{tabular}{ccc}
 \multirow{2}{*}{\begin{tabular}{c} $k$-NN \\[-1 mm] $\|\;\|_\infty$ \end{tabular}} &
 \multirow{2}{*}{\begin{tabular}{c} $k$-NN \\[-1 mm] PLS \end{tabular}} &
 \multirow{2}{*}{\begin{tabular}{c} Nonpar. \\[-1 mm] plug-in \end{tabular}} \\
  \\ \hline
0.79 & 0.66 & 0.85 \\ \hline
\end{tabular}
\caption{Proportion of correctly classified for the transformed cell data.}
\label{CellsTable2}
\end{center}
\end{table} 

\end{document}